%% file: main.tex
\pdfoutput=1

\documentclass[11pt]{article}
\usepackage{EMNLP2022}

\usepackage{times}
\usepackage{latexsym}

\usepackage[T1]{fontenc}

\usepackage[utf8]{inputenc}

\usepackage{microtype}

\usepackage{booktabs}
\usepackage{amsmath,amsfonts,amsthm,amssymb,mathtools}
\usepackage{bbm}
\usepackage{setspace}
\usepackage{fancyhdr}
\usepackage{mathrsfs}
\usepackage{lastpage}
\usepackage{longtable}
\usepackage{multirow}
\usepackage{extramarks}
\usepackage{chngpage}
\usepackage{siunitx}
\usepackage{soul,color}
\usepackage{enumerate}
\usepackage{listings}
\usepackage{verbatimbox}
\usepackage{graphicx,float,wrapfig}
\usepackage{tikz}
\usepackage{tkz-euclide}
\usetikzlibrary{calc,automata,positioning}
\usepackage{xcolor,colortbl}
\usetikzlibrary{shapes}

\usepackage{caption}
\usepackage{subcaption}

\usepackage{tikz}
\usepackage{tikz-qtree}
\usepackage{tikz-qtree-compat}
\usepackage{tikz-dependency}

\usepackage{forest}
\useforestlibrary{linguistics}
\usetikzlibrary{decorations,decorations.text}
\usetikzlibrary{backgrounds,fit}

\usepackage{ragged2e}

\usepackage{natbib}

\usepackage[normalem]{ulem}

\usepackage[ruled,vlined]{algorithm2e}
\SetKwInOut{Parameter}{parameter}

\input{math_commands}


\input{header}

\title{Finding Dataset Shortcuts with Grammar Induction}
\author{%
  Dan Friedman, Alexander Wettig, Danqi Chen \\
    Department of Computer Science, Princeton University\\
    \texttt{\{dfriedman,awettig,danqic\}@cs.princeton.edu} \\
}

\begin{document}
\maketitle

\begin{abstract}
Many NLP datasets have been found to contain shortcuts: simple decision rules that achieve surprisingly high accuracy.
However, it is difficult to discover shortcuts automatically.
Prior work on automatic shortcut detection has focused on enumerating features like unigrams or bigrams, which can find only low-level shortcuts, or relied on post-hoc model interpretability methods like saliency maps, which reveal qualitative patterns without a clear statistical interpretation.
In this work, we propose to use probabilistic grammars to characterize and discover shortcuts in NLP datasets.
Specifically, we use a context-free grammar to model patterns in sentence classification datasets and use a synchronous context-free grammar to model datasets involving sentence pairs.
The resulting grammars reveal interesting shortcut features in a number of datasets, including both simple and high-level features, and automatically identify groups of test examples on which conventional classifiers fail.
Finally, we show that the features we discover can be used to generate diagnostic contrast examples and incorporated into standard robust optimization methods to improve worst-group accuracy.\footnote{Our code for inducing grammars and finding dataset shortcuts is available at \href{https://github.com/princeton-nlp/ShortcutGrammar}{https://github.com/princeton-nlp/ShortcutGrammar}.}

\end{abstract}

\input{sections/intro}

\input{sections/background}

\input{sections/method}

\input{sections/analysis}

\input{sections/experiments}

\input{sections/related}

\section{Conclusion}
We have developed an approach for automatically finding dataset shortcuts by inducing dataset-specific grammars. We demonstrated that it reveals interesting shortcut features in four classification datasets and can be used as a diagnostic tool to identify categories of examples on which classifiers are more likely to fail.
Future work will explore approximate inference methods to scale this approach to datasets with longer sequences, and extensions to more expressive grammar formalisms.

\section*{Acknowledgements}
We thank the members of the
Princeton NLP group and the anonymous reviewers
for their valuable comments and feedback.

\newpage
\section*{Limitations}
Our method has several limitations that raise interesting challenges for future work.

\paragraph{Grammar scalability}
First is the scalability of the synchronous parsing algorithm, which limits us in practice from applying this approach to datasets with long sentence pairs. 
 One direction for future work is to explore approximate inference techniques that will allow these methods to scale to large datasets with longer sequences and use fewer computational resources.

\paragraph{Grammar expressiveness}
Second is the expressiveness of the grammar, which assumes that pairs of sentences can be modeled with a single parse tree.
This assumption works well for relatively simple datasets like SNLI, but is less reasonable for sentences that have very different syntactic structure.
An interesting direction for future work is to explore more expressive grammar formalisms, such as synchronous tree substitution grammars~\citep{shieber1990synchronous}.

\paragraph{Comparing shortcut finding methods} There are many existing approaches to finding shortcuts as we discussed in the paper, but it is difficult to have an apples-to-apples comparison. In particular, attribution-based methods~\citep{han2020explaining,pezeshkpour2022combining} do not provide explicit feature representation of shortcuts, and so it is difficult to say whether these methods are capable of identifying the same patterns. We have shown qualitative comparisons of the kinds of features provided by different methods but leave the question of better quantitative evaluation to future work.

\paragraph{Applications to other languages and tasks}
We only apply our approach to four English-language datasets.
In particular, the sentence pair datasets we consider are based on sentence similarity judgments, where synchronous grammars are a good choice.
We hope to apply the approach to other tasks and other languages in the future and explore formalisms that are better suited to modeling more diverse relationships between strings.

\paragraph{Out-of-domain generalization}
We conducted one set of experiments with robust optimization and showed some improvements on in-domain minority examples.
In future work, we are also interested in exploring whether our shortcut-finding methods can be useful for improving performance in generalization to other out-of-domain distributions.

\bibliographystyle{acl_natbib}
\bibliography{ref}

\appendix

\input{sections/appendix.tex}

\end{document}

%% file: math_commands.tex
\usepackage{amsmath,amsfonts,bm}

\def\1{\bm{1}}

\def\sx{{\bm{x}}}

\makeatletter
\newcommand{\ve}{\@ifnextchar\bgroup{\velong}{{\bm{e}}}}
\newcommand{\velong}[1]{{\bm{#1}}}
\makeatother

\def\ve{{\mathbf{e}}}

\def\vu{{\mathbf{u}}}
\def\vv{{\mathbf{v}}}

\DeclareMathAlphabet{\mathsfit}{\encodingdefault}{\sfdefault}{m}{sl}
\SetMathAlphabet{\mathsfit}{bold}{\encodingdefault}{\sfdefault}{bx}{n}

\def\gD{{\mathcal{D}}}

\def\gG{{\mathcal{G}}}

\def\gN{{\mathcal{N}}}

\def\gP{{\mathcal{P}}}

\def\gR{{\mathcal{R}}}

\def\gV{{\mathcal{V}}}

\def\gX{{\mathcal{X}}}
\def\gY{{\mathcal{Y}}}

\newcommand{\R}{\mathbb{R}}

\DeclareMathOperator*{\argmax}{arg\,max}

%% file: header.tex
\definecolor{patterncolor}{rgb}{0.94, 0.97, 1.0}
\definecolor{poscolor}{rgb}{0.97, 1.0, 0.94}
\definecolor{negcolor}{rgb}{1.0, 0.97, 0.94}
\definecolor{neutcolor}{rgb}{0.98, 0.96, 1.0}
\definecolor{nodehighlightcolor}{rgb}{1.0, 1.0, 1.0}
\definecolor{edgehighlightcolor}{rgb}{0.0, 0.3, 0.8}
\definecolor{saliencycolor}{rgb}{0.7, 0.95, 0.85}

\forestset{
    default preamble={
        for tree={
            highlight/.style={
                fill=nodehighlightcolor,
                very thick,
                draw={edgehighlightcolor},
                for descendants={
                    edge={very thick, edgehighlightcolor},
                    if n children=0{
                    }{}
                }
            },            
            thinnerhighlight/.style={
                fill=nodehighlightcolor,
                very thick,
                font=\fontsize{9}{11}\selectfont,
                draw={edgehighlightcolor},
                for descendants={
                    edge={thick, edgehighlightcolor},
                    if n children=0{
                    }{}
                }
            },
            neg/.style={
                fill=negcolor
            },
            if n children=0{
              rectangle,
              rounded corners=5pt,
              fill=patterncolor,
              font=\small,
              tier=terminal,
              align=center
            }{
                circle,
                draw,
                inner sep=0,
                minimum height=12pt,
            },
        }
    }
}

\definecolor{aliceblue}{rgb}{0.94, 0.97, 1.0}
\definecolor{babyblue}{rgb}{0.54, 0.81, 0.94}
\definecolor{bisque}{rgb}{1.0, 0.89, 0.77}
\definecolor{green(munsell)}{rgb}{0.0, 0.66, 0.47}
\definecolor{tiffanyblue}{rgb}{0.04, 0.73, 0.71}
\definecolor{titaniumyellow}{rgb}{0.93, 0.9, 0.0}
\definecolor{peach-yellow}{rgb}{0.98, 0.87, 0.68}
\definecolor{blond}{rgb}{0.98, 0.94, 0.75}
\definecolor{lightmauve}{rgb}{0.86, 0.82, 1.0}

\newcommand\ti[1]{\textit{#1}}

\newcommand\tf[1]{\textbf{#1}}

\newcommand*\rootlabel[1]{\tikz[baseline=(char.base)]{
        \node[shape=circle,draw,inner sep=0pt, outer sep=0pt,minimum height=12pt] (char) {#1};}}

\newcommand*\textex[1]{\tikz[baseline=(char.base)]{
        \node[shape=rectangle, rounded corners=5pt, inner xsep=4pt, inner ysep=0pt, outer sep=0pt, fill=patterncolor, minimum height=10pt] (char) {#1};}}
\newcommand*\textexneut[1]{\tikz[baseline=(char.base)]{
        \node[shape=rectangle, rounded corners=5pt, inner xsep=4pt, inner ysep=0pt, outer sep=0pt, fill=neutcolor, minimum height=10pt] (char) {#1};}}
\newcommand*\textexpos[1]{\tikz[baseline=(char.base)]{
        \node[shape=rectangle, rounded corners=5pt, inner xsep=4pt, inner ysep=0pt, outer sep=0pt, fill=poscolor, minimum height=10pt] (char) {#1};}}
\newcommand*\textexneg[1]{\tikz[baseline=(char.base)]{
        \node[shape=rectangle, rounded corners=5pt, inner xsep=4pt, inner ysep=0pt, outer sep=0pt, fill=negcolor, minimum height=10pt] (char) {#1};}}

\renewcommand{\paragraph}[1]{\vspace{0.2cm}\noindent\textbf{#1}}

%% file: sections/intro.tex
\input{figures/grammar_example.tex}

\vspace{-1em}
\section{Introduction}
\label{sec:introduction}
Many NLP datasets have been found to contain shortcuts: simple decision rules that achieve surprisingly high accuracy.
For example, it is possible to get high classification accuracy on
paraphrase identification datasets by predicting that sentences with
many common words are paraphrases of each other~\cite{zhang2019paws}.
Such classifiers are said to be ``right for the wrong
reason''~\citep{mccoy2019right}.
Shortcuts are a problem if they do not generalize to the intended test distribution~\citep{geirhos2020shortcut}.
For example, paraphrase identification models might misclassify
non-paraphrases that have many overlapping words. %

Shortcuts have been reported in many established
datasets~\citep[e.g.,][]{mccoy2019right,gururangan2018annotation,niven2019probing,schuster2019towards},
as a consequence of \ti{annotation artifacts} or so-called \ti{spurious correlations}.
Typically, these discoveries are the result of human intuition about possible patterns in a particular dataset.
Our goal in this paper is to discover shortcuts automatically.
If we can identify shortcuts, we can identify categories of examples on which conventional classifiers will fail, and try to mitigate these weaknesses by collecting more training data or using robust optimization algorithms.

The main challenge to automatically identifying shortcuts is to develop a formal framework for describing patterns in language data that can capture both simple and high-level features and that makes it possible to search for these patterns efficiently.
Prior work has addressed only simple features like unigrams and bigrams~\citep{wang2020identifying,wang2022identifying,gardner2021competency}, and it is difficult to extend this approach to more sophisticated patterns, like lexical overlap, without knowing the pattern in advance.
Other model-based approaches use black-box interpretability methods, by
using gradient-based
techniques to identify tokens or training instances that influence a
particular
prediction~\citep{han2020explaining,pezeshkpour2022combining,bastings2021will}.
These methods offer local, qualitative hints about the decision of a classifier
on a particular test instance, but do not identify dataset-level features nor provide a way of measuring the strength of correlation.

Our approach is to use grammar induction to characterize and discover
shortcut features (\S\ref{sec:method}). Probabilistic grammars provide a principled
framework for describing patterns in natural language, allowing us to
formally model both simple features and more
sophisticated patterns. They admit tractable
search algorithms and provide a natural way to measure statistics
about the correlation between text patterns and labels.
Grammars also offer a mechanism for identifying contrastive features,
templates that appear in similar contexts across classes, but take on
different values,  which can be used to construct diagnostic test examples (see examples in Figure~\ref{fig:grammar_example}).

In this work, we focus on both single-sentence (e.g., sentiment analysis) and sentence-pair classification datasets (e.g., NLI, paraphrase detection).
While we can use context-free grammars to model features in sentences,  sentence-pair datasets present a particular challenge, as it is difficult to enumerate interactions between the pair of sentences. We propose to use synchronous context-free grammars, which formally model insertion, deletion, and alignment.
We find that we can extract meaningful, dataset-specific structures that describe latent patterns characterizing these classification tasks.

We apply our approach to four classification datasets: IMDb, SUBJ, SNLI and QQP. 
After illustrating the shortcut features  (\S\ref{sec:finding_shortcuts}), we explore whether state-of-the-art classifiers exploit these shortcuts by identifying minority examples in the datasets and then generating contrastive examples (\S\ref{sec:model_analysis}).
Then we demonstrate that these features can be used in robust optimization algorithms to improve generalization (\S\ref{sec:remedying_shortcuts}).
Finally, we compare this approach with model-based interpretability methods and n-gram features, which do not explicitly model syntax (\S\ref{sec:comparison_others}). Overall, we find that grammar induction provides a flexible and expressive representation for modeling features in NLP datasets.

%% file: figures/grammar_example.tex
\begin{figure*}

\centering
\begin{subfigure}{.45\linewidth}
  \centering
  \vspace{0.5cm}
  \resizebox{\linewidth}{!}{%
    \begin{forest} 
        [{1}
          [{14},
            [{14},highlight,tikz={\node [draw, dotted, thick, rounded corners=5pt, inner xsep=2pt, edgehighlightcolor, fit=()(!1)(!ll)]{};},
            [{41} [A\\/\\A] ] [{32} [girl\\/\\kid] ] ]
            [{10}
              [{16} [{75}   [in\\/\\ $\epsilon$]] [{94} [a\\/\\ $\epsilon$] ] ]
              [{11} [{70}
              [green\\ /\\ $\epsilon$]] [{70} [shirt\\ /\\ $\epsilon$]] ] ] ]
          [{31}
            [{77},highlight,tikz={\node [draw, dotted, thick, rounded corners=5pt, inner xsep=2pt, edgehighlightcolor, fit=()(!1)]{};},
                [jumps\\/\\sits] ]
            [{2},highlight,tikz={\node [draw, dotted, thick, rounded corners=5pt, inner xsep=2pt, edgehighlightcolor, fit=()(!11)(!ll)]{};},
            [{0} [{59} [in\\/\\in] ] [{38} [the\\/\\a] ] ] [{83} [air\\/\\chair] ] ] ] ]
    \end{forest}
  }
  \label{fig:sub1}
\end{subfigure}\hfill%
\begin{subfigure}{.54\linewidth}
  \centering
  \scriptsize
  \begin{tabular}{ccc}
        \toprule 
       \begin{forest} 
        [{14},thinnerhighlight,
            [A girl/A kid,roof]]
        \end{forest} &
       \begin{forest} 
            [{77},thinnerhighlight,
            [jumps/sits]]
        \end{forest} &
        \begin{forest} 
        [{2},thinnerhighlight,
            [in the air/in a chair,roof]]
        \end{forest} \\
        \midrule
       
        \multicolumn{3}{c}{Top subtrees for \ti{entailment}} \\
        \midrule
        \textexpos{A man/A person} & \textexpos{walking/walking} & \textexpos{in the grass/outside} \\ 
        \textexpos{A man/A human} & \textexpos{walk/walking} &  \textexpos{down the street/outside} \\
        \midrule 
        \multicolumn{3}{c}{Top subtrees for \ti{contradiction}} \\
        \midrule
        \textexneg{A man/A woman} & \textexneg{$\epsilon$/sitting} & \textexneg{at night/during the day} \\ 
        \textexneg{A woman/A man} & \textexneg{stand/sit} &  \textexneg{in the air/down} \\
        \midrule
        \multicolumn{3}{c}{Top subtrees for \ti{neutral}} \\
        \midrule
       \textexneut{A man/A tall human} & \textexneut{$\epsilon$/competes} & \textexneut{down the street/home} \\ 
        \textexneut{A man/An old man} & \textexneut{running/running} &  \textexneut{in the sand/on the beach} \\
        \bottomrule
  \end{tabular}
  \label{fig:sub2}
\end{subfigure}
\caption{
Left: The most likely parse tree for an example from the SNLI validation set according to our synchronous grammar.
The numbered nodes index non-terminal symbols and  $\epsilon$ denotes an empty symbol.
We highlight the subtrees that provide the strongest evidence in favor of the label \ti{contradiction}, and show alternative spans generated by these non-terminals after conditioning on the class labels (Right). 
}
\label{fig:grammar_example}
\end{figure*}

%% file: sections/background.tex
\section{Method}
\label{sec:method}

\subsection{Overview}
\label{sec:background}

%% file: sections/method.tex
We focus on text classification datasets $\gD \subseteq \gX \times \gY$, where $y \in \gY$ is a categorical label and $x \in \gX$ is either a sentence or a pair of sentences, each consisting of a sequence of words from a discrete vocabulary $\gV$.
When $x$ is a sentence pair, we will write $x = (x^a, x^b)$ and refer to $x^a$ and $x^b$ as the source and target sentences, respectively.

We aim to automatically extract a description of the features that characterize the relationship between $x$ and $y$,
and our key idea is to define the features in terms of a \ti{dataset-specific grammar}.
Compared to a grammar extracted from a standard treebank, the grammars we induce serve as interpretable models for the distribution of sentences in the dataset.
Our approach consists of two steps:
\begin{enumerate}
 \setlength{\itemsep}{-0.5pt}
  \item \tf{Grammar induction:} First, we induce a grammar for (unlabeled) training instances $x_1, \ldots, x_N$ and get the maximum likelihood trees $t_1, \ldots, t_N$.
    \item \tf{Finding features:} We define features in terms of subtrees in the grammar, which describe patterns in the input sentences, and we search for features that have high mutual information with the class labels.
\end{enumerate}
\vspace{-0.5em}
We induce one shared grammar for all classes rather than incorporating labels during grammar induction, so that the non-terminal symbols have a consistent meaning across classes.
This facilitates finding contrastive features.
For example, in Figure~\ref{fig:grammar_example} (right), the nonterminal symbol \rootlabel{77} always generates pairs of verbs, but the distribution changes according to the class label: \rootlabel{77} is more likely to generate \textex{walk/walking} when the class label is \ti{entailment} and more likely to generate \textex{stand/sits} when the class label is \ti{contradiction}.

\subsection{Grammar Induction}
\label{sec:grammar_features}
In this section, we describe the two grammars we use and the training procedure.

\paragraph{Context-free grammar}
A context-free grammar (CFG) consists of an inventory of terminal symbols $\gV$ (words), non-terminal symbols $\gN$, and production rules of the form $\alpha \to \beta \in \gR$, where $\alpha \in \gN$ and $\beta \in (\gV \cup \gN)^*$.
A probabilistic CFG (PCFG) defines a distribution over trees by assigning every non-terminal symbol a categorical distribution over production rules, with the probability of a tree defined as the product of the probability of the production rules used to generate it.
A PCFG defines a distribution over sequences of words $x \in \gV^*$ by marginalizing over all trees which generate the sentence $x$ (denoted by $\mathrm{yield}(t)$):\[
  p(x) = \sum_{t: \mathrm{yield}(t) = x} p(t).
\vspace{-0.5em}
\]

\paragraph{Synchronous grammar}
Many NLP shortcuts have been found in datasets involving pairs of sentences $x = (x^a, x^b)$. We model patterns in these datasets using a
Synchronous PCFG~\citep[SCFG;
][]{lewis1968syntax,wu1997stochastic}, a grammar for defining probability distributions over pairs of sequences.
An SCFG assumes that both sequences were generated from a single context-free parse tree, whose terminal symbols have the form $w^a/w^b$, where $w^a$ and $w^b$ are either words in $x^a$ or $x^b$ respectively, or an empty symbol, denoted by $\epsilon$, which represents a null alignment (Figure~\ref{fig:grammar_example}).
SCFG productions can also be thought of as translation rules: the emission $w^a/w^b$ represents a substitution---replacing word $w^a$ with $w^b$---and null alignments represent insertion or deletion.
An SCFG makes strong simplifying assumptions about the possible relationships between sentences, but, as we will show, it is still capable of modeling interesting, hierarchical structure in real-world datasets.

\paragraph{Parameterization and training}
The parameters of grammar consist of a vector $\theta$ with one entry $\theta_{\alpha \to \beta}$ for every rule $\alpha \to \beta \in \gR$.
Following~\citet{kim2019compound}, we parameterize the grammars using a neural network.
We use the neural CFG parameterization from~\citet{kim2019compound} and develop 
a similar parameterization for our SCFG, with extensions for the terminal production rules.
We defer the full details to the appendix (Section~\ref{appendix:parameterization}).

Given a training set $\gD = \{(x_i, y_i)\}_{i=1}^N$ and a grammar
$\gG = (\gV, \gN, \gR)$, we find a maximum likelihood estimate $\theta^*$ by maximizing the marginal likelihood of the (unlabeled) training sentences:
\begin{align*}
  \theta^* &= \argmax_{\theta} \sum_{i=1}^N \log p(x_i \mid \gG, \theta).
\end{align*}
We optimize the parameters using gradient descent.
After training, we calculate the maximum likelihood trees $t_1, \ldots, t_N$ for the training data, and use these trees as the basis of further analysis.

\paragraph{Complexity}
Training and parsing require enumerating the trees consistent with the input, which is calculated with the inside algorithm for CFGs~\citep{lari1990estimation} and the bitext inside algorithm for SCFGs~\citep{wu1997stochastic}.
The inside algorithm has space and time complexity of $O(|x|^3|\mathcal{G}|)$ and the bitext inside algorithm has space and time complexity of $O(|x^a|^3|x^b|^3|\mathcal{G}|)$, where $|\mathcal{G}|$ is a grammar constant determined by the number of rules in the grammar.
We use a vectorized GPU implementation of the inside algorithm provided by Torch-Struct~\citep{rush2020torch} and we implement a vectorized version of the bitext inside algorithm.
The cost of the bitext parsing algorithm imposes a practical limitation on the length of the sentences we consider (Section~\ref{sec:experimental_setup}). We also discuss possible efficient approximations in the Limitations section, which we leave as an avenue for future work.

\subsection{Finding Features}
\label{sec:finding_features}
The tree-annotated corpus provides a structured representation of the dataset that we can now query to find discriminative patterns at various levels of abstraction.
We follow a simple procedure for finding dataset-level patterns using our tree annotations.

First, given a set of trees $t_1, \ldots, t_N$ with class labels $y_1, \ldots, y_N$, we extract the set of complete subtrees, which are subtrees whose leaves are all terminal symbols. There is at most one unique subtree for each non-terminal node in each tree, so the number of complete subtrees is roughly on the order of the number of words in the training data.

Next, we calculate the \ti{mutual information} between each subtree and the class labels.
We treat each subtree $s$ as a boolean-valued feature function on trees, $\phi_s(t) = \mathbbm{1}[s \in t]$.
Let $Z_s$ be a random variable denoting the output of $\phi_s$ and let $Y$ be a random variable over $\gY$.
The mutual information is defined as:
\begin{align*}
    I(Z_s; Y) = \sum_{z_s \in \{0, 1\}} \sum_{y \in \gY} p(y, z_s) \log \frac{p(y, z_s)}{p(y) p(z_s)}.
\end{align*}
 We estimate the mutual information between $Z_s$ and $Y$ using $\hat{p}(y, z_s) \propto 1 + \sum_{i=1}^N \mathbbm{1}[y_i = y \wedge \phi_s(t_i) = z_s]$. 
Mutual information measures the expected amount of information we learn about $Y$ by learning the value of $\phi_s$.
While we use mutual information in this paper, we could also score the features using other feature-importance metrics, such as z-score~\citep{gardner2021competency}.

To visualize the most discriminative patterns, we group the highest-ranked subtrees according to their root label and majority class label.
Let $S(\alpha, y)$ denote the set of subtrees with root label $\alpha$ and majority class label $y$. We define a composite feature, $Z_{\alpha, y} = \bigvee_{s \in S(\alpha, y)} Z_s$ as the union of features in $S(\alpha, y)$.
The result of this procedure is a concise list of class-conditional non-terminal features, which we can inspect to identify the patterns that are broadly discriminative across the dataset.

%% file: sections/analysis.tex
\section{Finding Shortcuts}
\label{sec:finding_shortcuts} 
\subsection{Experimental Setup}
\label{sec:experimental_setup}
\paragraph{Datasets}
We apply our approach to two single-sentence and two sentence-pair classification datasets.
\tf{IMDb}~\citep{maas2011learning} is a binary sentiment analysis dataset consisting of paragraph-length movie reviews.
\tf{SUBJ}~\citep{pang2004sentimental} is a subjectivity classification dataset, containing sentences labeled as either \ti{subjective} or \ti{objective}.
\tf{SNLI}~\citep{bowman2015large} is a three-way classification task; given two sentences, $x^a$ and $x^b$, the objective is to determine whether $x^a$  \ti{entails} $x^b$, \ti{contradicts} $x^b$, or is \ti{neutral}.
\tf{QQP}~\citep{iyer2017first} consists of pairs of questions from quora.com labeled as being either \ti{paraphrases} or \ti{not paraphrases}.

For all experiments, we fix the size of the grammar to be 96 non-terminals symbols, divided into 32 internal non-terminal symbols and 64 pre-terminal symbols, similar to~\citet{kim2019compound},
and use a lowercase word-level tokenizer with a maximum vocabulary size of 20,000 words.
For the sentence-pair datasets, we randomly sample 65K/16K class-balanced sets of training/validation examples that fit within the length limit imposed by the bitext inside-outside algorithm ($|x^a| \times |x^b| < 225$).
This length limit covers approximately 80\% of SNLI and 70\% of QQP. 
For IMDb, we split the movie reviews into sentences for the purposes of training the PCFG, but compute feature statistics using the full reviews.
More implementation details are in Appendix~\ref{appendix:training_details}.

\input{tables/imdb_features}

\subsection{A Look at the Top Features}
In this section, we qualitatively explore some of the dataset-level shortcuts we find.
Our procedure for each dataset is the same: given a set of trees, we enumerate all complete subtrees and sort them by mutual information, treating each subtree as a binary indicator variable. Then we group the subtree features by root label and majority class label and inspect the most informative groups of subtrees.
The majority class label for a feature $Z$ is defined as the most common class label among training examples for which $Z = 1$. For each feature $Z$, we report the number of training examples for which $Z = 1$ (\tf{Count}) and the percentage of these having the majority class label (\tf{\% Majority}).
We present the most interesting results here and include extended results in Appendix~\ref{appendix:additional_features}.

\paragraph{IMDb}
Not surprisingly, we find that the most informative features in IMDb include adjectives, adverbs, and nouns with high sentiment valence. We highlight some of the more interesting patterns in Table~\ref{tab:imdb_features}. For example, we discover that negative reviews are almost three times as likely as positive reviews to mention the length of the film (node \rootlabel{8}), and we find that the grammar has learned a clear category corresponding to names (node \rootlabel{5}).

We also confirm that the grammar recovers a known shortcut in IMDb, numerical ratings~\citep{ross2021explaining,pezeshkpour2022combining}.
We find that a single non-terminal (node \rootlabel{29}) is responsible for generating ratings, including both simple, numerical ratings, which have been documented in earlier work, as well as letter grades and ratings on a star system (Table~\ref{tab:imdb_ratings}).
\input{tables/imdb_ratings}

\input{tables/subj_features}

\paragraph{SUBJ}
In the SUBJ dataset (Table~\ref{tab:subj_features}), the most informative features
reflect how this dataset was constructed~\citep{pang2004sentimental}: the \ti{subjective} class consists of movie reviews from Rotten Tomatoes and the \ti{objective} class consists of movie summaries from IMDb.
A similar observation was made by~\citet{zhong2022summarizing}, who trained a neural network to generate natural language descriptions of the differences between text distributions.
Our method points us to the same conclusion, but by modeling the statistics of the dataset rather than querying a black-box neural network.

\input{tables/snli_features_v2}

\paragraph{SNLI}
Now we consider the synchronous grammar features for SNLI (Table~\ref{tab:snli_features}).
Prior work has documented the presence of hypothesis-only shortcuts in SNLI as well as shallow cross-sentence features like lexical overlap~\citep{mccoy2019right,gururangan2018annotation,poliak2018hypothesis}. 
The SCFG features reveal a number of more sophisticated patterns and clusters them in clear categories.
The \ti{contradiction} class has the most highly discriminative shortcut features, which mainly involve replacing a subject or verb word with a direct \tf{antonym}.
The most informative \ti{neutral} features involve \tf{additions}, such as adding adjectives or prepositional phrases.
The highest scoring \ti{entailment} examples include \tf{hypernyms}, such as changing ``man'' to ``human''. 
These features explicitly model the alignment between grammatical roles, as well as insertion and deletion, giving a high-level view of some of the common strategies employed by crowd-workers in creating this dataset.

Additional, higher-level features are presented in Appendix Table~\ref{tab:snli_nonterminals}, which lists the internal production rules with the highest mutual information.
In general, these features are less discriminative than lexicalized features, but they describe more abstract properties, such as removing prepositional phrases from $x^a$ to create an entailment.

\input{tables/qqp_features_v2}
\paragraph{QQP}
The highest scoring features in QQP are listed in Table~\ref{tab:qqp_features}.
The best known shortcut in QQP is lexical overlap, which is more likely to be high between paraphrases, and the SCFG features echo this fact: most of the highest ranking \ti{paraphrase} features are pairs of aligned words or phrases, and the highest ranking \ti{no paraphrase} features are function words that have no alignment in the corresponding question.
Other prominent features involve changes to the question structure, as well as a number of specific topics that are surprisingly prevalent in the dataset and provide strong evidence that a pair of questions are paraphrases, including open-ended discussion topics such as New Year's resolutions, World War 3, and the 2016 presidential election, as well as lifestyle advice about how to make money or lose weight.

\section{Do Models Exploit These Shortcuts?}
\label{sec:model_analysis}
In this section, we explore how the shortcuts we have discovered affect the generalization behavior of conventional classifiers that are trained on the same data.
We train BERT-base~\cite{devlin2019bert} and RoBERTa-base~\cite{liu2019roberta} classifiers on our SNLI and QQP training splits and examine the performance by finding minority groups of counter-examples in the dataset, and then by designing contrastive examples using the grammar.

\input{tables/combined_errors.tex}

\paragraph{Observational counter-examples}
First, for each shortcut feature $Z$, let $y^{Z=1}$ denote the majority label for training instances  that contain the shortcut feature ($Z=1$). We take the validation
examples $x_i, y_i$ for which $Z = 1$ and partition them into \ti{supporting examples} and \ti{counter-examples} %
according to whether or not $y_i = y^{Z=1}$.
We report the accuracy of the BERT and RoBERTa models on supporting and counter-examples in Table~\ref{tab:combined_errors}.
Both models consistently perform higher than average on the supporting examples and much worse on the counter-examples, indicating that the grammar feature are at least correlated with the features used by these classifier.
This trend is consistent for all the features, perhaps suggesting that these models exploit every available feature to some extent.

\paragraph{Creating contrastive examples}
In the previous section, we established that there is a correlation between shortcut features and BERT error rates using counter-examples that appear in the training data.
In this section, we generate controlled contrasting examples to test specific hypotheses about what function the model has learned.
Our procedure in this section is similar to counter-factual data augmentation~\citep{kaushik2020learning} and contrast sets~\citep{gardner2020evaluating}. The difference is that our edits are based on explicit feature representations derived from the dataset, allowing us to better control for confounding features.

\input{tables/snli_contrasting_example_results}

We focus on the three patterns we highlighted in SNLI: simple hypernyms, antonyms, and additions.
For each feature $Z$ with majority label $y^{Z=1}$, we design a rule-based edit that will select and perturb existing validation instances $(x, y)$ to obtain instances $(x^*, y^*)$ for which $Z = 1$ but $y^* \neq y^{Z=1}$.
\tf{Hypernyms:} We select \ti{neutral} or \ti{contradiction} examples that have an aligned subject word (e.g. \textex{man/man}), replace the hypothesis subject word with a hypernym, and expect the label to stay the same. \tf{Antonyms:} Pick \ti{entailment} or \ti{neutral} examples with an adjective modifying the subject noun, add an antonym adjective to an object noun, and expect the label to become \ti{neutral}. \tf{Add adjective:} Pick \ti{contradiction} examples, add an adjective modifying the subject noun, and expect the label to stay the same.
In each case, we use the grammar to identify the set of possible edits.
That is, we select antonyms like \textex{white/black} or \textex{small/large} and add adjectives like \textex{$\epsilon$/tall} and \textex{$\epsilon$/sad}. %

For each validation example $(x_i, y_i)$, we create a contrast set $\mathcal{S}_i$ consisting of one or more perturbations of $x_i$. For example, if $x_i$ contained \textex{man/man}, $\mathcal{S}_i$ will include examples containing \textex{man/person} and \textex{man/human}.
We report the \tf{error rate}, defined as the percentage of the contrast sets $\mathcal{S}_i$ for which the model predicts $y^{Z=1}$ for any $x^* \in \mathcal{S}_i$, restricted to sets such that the model classified the original $(x_i, y_i)$ correctly.

The results of this experiment are in Table~\ref{tab:snli_contrasting_example_results}, along with an example of each edit.
On each test set, we find many perturbations that lead the model to change its prediction. The model performs worst on the Antonyms test set, suggesting that the presence of contradicting adjectives may be a strong signal to the model, regardless of whether the adjectives are attached to the same entity.

%% file: tables/imdb_features.tex
\begin{table*}[ht!]
\centering
\resizebox{\linewidth}{!}{%
\begin{tabular}{l c l p{0.8\linewidth} r r}
\toprule
 & \tf{Root} & \tf{Description} & \tf{Patterns} & \tf{Count} & \% \tf{Majority} \\
\midrule
N & \rootlabel{5} & Negative actors & \textexneg{ed wood}, \textexneg{steven seagal}, \textexneg{uwe boll}, \textexneg{van damme}, \textexneg{tom savini} & 174 & 95.5\\
  & \rootlabel{29} & Negative ratings & 
\textexneg{4 / 10}, \textexneg{3 / 10}, \textexneg{1 / 10}, \textexneg{2 / 10}, \textexneg{1 / 2 from * * * *} & 429 & 96.8\\
 & \rootlabel{8} & Negative durations  & \textexneg{30 minutes}, \textexneg{10 minutes}, \textexneg{five minutes}, \textexneg{90 minutes}, \textexneg{2 hours} & 1,412	& 76.7\\
  \midrule
P & \rootlabel{5} & Positive actors & \textexpos{walter matthau}, \textexpos{jon voight}, \textexpos{james stewart}, \textexpos{william powell}, \textexpos{philo vance} & 751 & 88.6\\
 & \rootlabel{29} & Positive ratings & \textexpos{10 / 10}, \textexpos{8 / 10}, \textexpos{7 / 10}, \textexpos{highly recommended .}, \textexpos{9 / 10} & 486 & 98.8\\
 & \rootlabel{8} & Positive durations & \textexpos{many years}%
 & 95 & 69.1\\
\bottomrule
\end{tabular}
}
\caption{
\label{tab:imdb_features}
Six high-scoring features in IMDb, grouped by majority class (N: Negative, P: Positive) and showing at most five spans per row, with our own descriptions of the pattern reflected in each row.%
}
\end{table*}

%% file: tables/imdb_ratings.tex
\begin{table}[t]
\centering
\resizebox{\columnwidth}{!}{%
\begin{tabular}{l p{0.7\linewidth} r r}
\toprule
& \tf{Patterns} & \tf{Count} & \% \tf{Majority} \\
  \midrule
N & {\RaggedLeft \textexneg{3 out of 10 .}, 
\textexneg{4 out of 10 .}, 
\textexneg{1 out of 10 .}} & 41 & 100.0 \\
& {\RaggedLeft \textexneg{my grade : d}, \textexneg{my grade : f}, \textexneg{my grade : c}} & 33 & 100.0\\
& \textexneg{1 / 2 from * * * *} & 44 & 93.2\\
  \midrule
P & {\RaggedLeft \textexpos{10 out of 10 .}, \textexpos{7 out of 10 .}, \textexpos{8 out of 10.}} & {51} & {100.0} \\
& \textexpos{my vote is eight .}, \textexpos{my vote is seven .} & 40 & 100.0\\
\bottomrule
\end{tabular}
}
\caption{
\label{tab:imdb_ratings}
Additional realizations of the ``movie rating'' pattern in IMDb (N: negative, P: positive).
All of these spans correspond to subtrees for root \rootlabel{29}.
}
\end{table}

%% file: tables/subj_features.tex
\begin{table}[t]
\centering
\resizebox{\columnwidth}{!}{%
\begin{tabular}{lc p{0.6\linewidth} cc}
\toprule
 & \tf{Root} & \tf{Patterns} & \tf{Count} & \% \tf{Majority} \\
\midrule
S & \rootlabel{27} & {\RaggedLeft \textexneg{a movie}, \textexneg{the film}, \textexneg{the movie}, \textexneg{this movie}} & 980 & 86.3 \\
& \rootlabel{3} & {\RaggedLeft \textexneg{comes off}, \textexneg{' s hard}, \textexneg{makes up}, \textexneg{' d expect}} & 460 & 87.4 \\
\midrule
O & \rootlabel{27} & {\RaggedLeft \textexpos{his life}, \textexpos{his wife}, \textexpos{his father}, \textexpos{his mother}} & 1,628 & 80.1\\
& \rootlabel{3} & {\RaggedLeft \textexpos{finds himself}, \textexpos{finds out}, \textexpos{falls in love}, \textexpos{is [UNK]} } & 205 & 85.5 \\
\bottomrule
\end{tabular}
}
\caption{
\label{tab:subj_features}
Four high-scoring features in SUBJ, filtering to subtrees with depth of at least 2, grouped by majority class (S: subjective, O: objective).  %
}
\end{table}

%% file: tables/snli_features_v2.tex
\begin{table*}[ht]
\centering
\resizebox{\linewidth}{!}{%
\begin{tabular}{l c l p{0.9\linewidth} r r}
\toprule
   & \tf{Root} & \tf{Description} & \tf{Patterns} & \tf{Count} & \% \tf{Majority} \\
\midrule
E & \rootlabel{44} & Copy verb & \textexpos{walking/walking}, \textexpos{running/running}, \textexpos{playing/playing}, \textexpos{sitting/sitting}, \textexpos{jumping/jumping} & 1,326 & 68.4\\
 & \rootlabel{14} & Subject phrase hypernym & \textexpos{a man/a person}, \textexpos{a man/a man}, \textexpos{a woman/a person}, \textexpos{ man/a man}, \textexpos{a man/a human} & 9,009 & 45.7\\
 & \rootlabel{4} & Expletive construction & \textexpos{a /there is}, \textexpos{$\epsilon$/there are}, \textexpos{two /there are}, \textexpos{a /there are}, \textexpos{$\epsilon$/there is} & 1,725 & 63.0\\
  \midrule
C & \rootlabel{32} & Subject antonym & \textexneg{man/woman}, \textexneg{woman/man}, \textexneg{boy/girl}, \textexneg{dog/cat}, \textexneg{girl/boy} & 1,235 & 91.1\\
 & \rootlabel{14} & Subject phrase antonym & \textexneg{a man/a woman}, \textexneg{a woman/a man}, \textexneg{a man/ nobody}, \textexneg{a boy/a girl}, \textexneg{a dog/a cat} & 1,351 & 82.5\\
 & \rootlabel{78} & Verb antonym & \textexneg{standing/sitting}, \textexneg{walking/sitting}, \textexneg{sitting/standing}, \textexneg{walking/running}, \textexneg{running/sitting} & 695 & 92.6\\
 & \rootlabel{41} & Definite article & \textexneg{a/the}, \textexneg{$\epsilon$/the} & 15,436 & 39.2\\
 & \rootlabel{85} & Adjective antonym &  \textexneg{black/white}, \textexneg{red/blue}, \textexneg{$\epsilon$/empty}, \textexneg{$\epsilon$/living}, \textexneg{white/black} & 560 & 76.9\\
  \midrule
N & \rootlabel{49} & Added function word & \textexneut{$\epsilon$/to}, \textexneut{ $\epsilon$/for}, \textexneut{ $\epsilon$/a}, \textexneut{ $\epsilon$/the}, \textexneut{ $\epsilon$/his} & 14,478 & 50.5 \\
 & \rootlabel{35} & Added object & \textexneut{$\epsilon$/[UNK]}, \textexneut{ $\epsilon$/work}, \textexneut{ $\epsilon$/get}, \textexneut{ $\epsilon$/friends}, \textexneut{ $\epsilon$/park} & 4557 & 59.8\\
 & \rootlabel{85} & Added adjective & \textexneut{$\epsilon$/tall}, \textexneut{$\epsilon$/sad}, \textexneut{$\epsilon$/[UNK]}, \textexneut{$\epsilon$/new}, \textexneut{$\epsilon$/big} & 1,945 & 72.1\\
 & \rootlabel{17} & Added PP phrase & \textexneut{$\epsilon$/to work}, \textexneut{$\epsilon$/to get}, \textexneut{$\epsilon$/to buy}, \textexneut{$\epsilon$/the park}, \textexneut{$\epsilon$/on vacation} & 1,411 & 71.4\\
\bottomrule
\end{tabular}
}
\caption{
\label{tab:snli_features}
Twelve of the highest scoring features in SNLI, grouped by majority class (E: {Entailment}, C: {Contradiction}, N: {Neutral}) with our own descriptions of the pattern reflected in each row.
$\epsilon$ stands for the empty string.
For each feature, we report the number of training examples and the percentage having the majority class label.
}
\end{table*}

%% file: tables/qqp_features_v2.tex
\begin{table*}[ht]
\centering
\resizebox{\linewidth}{!}{%
\begin{tabular}{l c l p{0.88\linewidth} r r}
\toprule
   & \tf{Root} & \tf{Description} & \tf{Patterns} & \tf{Count} & \% \tf{Majority} \\
\midrule
N & \rootlabel{70} & Additions & \textexneg{$\epsilon$/[UNK]}, \textexneg{$\epsilon$/in}, \textexneg{$\epsilon$/a}, \textexneg{$\epsilon$/-}, \textexneg{$\epsilon$/for} & 23,987 & 60.7\\
 & \rootlabel{49} & Deletions & \textexneg{[UNK]/$\epsilon$}, \textexneg{in/$\epsilon$}, \textexneg{a/$\epsilon$}, \textexneg{like/$\epsilon$}, \textexneg{of/$\epsilon$} & 21,299 & 61.6\\
 & \rootlabel{59} & Change question word & \textexneg{why/how}, \textexneg{why/what}, \textexneg{how/why}, \textexneg{why/can}, \textexneg{what/is} & 3,348 & 70.8 \\
  \midrule
P & \rootlabel{14} & How-to questions & {\RaggedLeft \textexpos{how can/how can}, \textexpos{how do/how can}, \textexpos{how can/how do}, \textexpos{how do/how do}, \textexpos{how can /what is the}} & 9,684 & 66.2\\
 & \rootlabel{25} & Discussion topics & {\RaggedLeft \textexpos{new year/new year}, \textexpos{world war/world war}, \textexpos{donald trump/donald trump}, \textexpos{hillary clinton/hillary clinton}, \textexpos{long distance/long distance}} & 2,179 & 82.9\\
 & \rootlabel{59} & Same question word & \textexpos{how/how}, \textexpos{why/why}, \textexpos{when/when}, \textexpos{how/what} & 17,264 & 60.3\\
\bottomrule
\end{tabular}
}
\caption{
\label{tab:qqp_features}
Six of the highest scoring features in QQP, grouped by majority class (N: non-paraphrase, P: paraphrase) with our own descriptions of the pattern reflected in each row.
$\epsilon$ stands for the empty string.
For each feature, we report the number of training examples and the percentage having the majority class label.
}
\end{table*}

%% file: tables/combined_errors.tex
\begin{table}[ht]
\centering
\resizebox{\linewidth}{!}{%
\begin{tabular}{l l  cccc}
\toprule
&   & \multicolumn{2}{c}{\tf{BERT}} & \multicolumn{2}{c}{\tf{RoBERTa}} \\
 &  & S & C & S & C \\
\midrule
\tf{SNLI} & \tf{Entailment} \\
& Copy verb & 98.6 & 83.6 & 98.6 & 83.6\\
& Subj. phrase hypernym & 92.5 & 80.3 & 92.4 & 83.8\\
& Expletive construction & 97.2 & 77.8 & 96.4 & 79.1\\
  \cmidrule{2-6}
  & \tf{Contradiction} \\
 & Subj. antonym & 96.9 & 61.5 & 97.3 & 76.9\\
 & Subj. phrase antonym  & 98.2 & 77.8 & 98.2 & 82.5\\
& Verb antonym  & 99.4 & 55.6 & 98.8 & 66.7\\
& Definite article & 86.2 & 82.2 & 88.9 & 82.9\\
& Adjective antonym & 91.7 & 71.4 & 95.8 & 78.6\\
  \cmidrule{2-6}
 & \tf{Neutral} \\
 & Added function word & 86.6 & 81.0 & 89.3 & 81.6\\
 & Added object  & 89.0 & 76.1 & 93.0 & 76.1\\
 & Added adjective  & 94.4 & 69.2 & 94.2 & 74.1\\
 & Added PP phrase & 92.9 & 77.0 & 95.0 & 77.0\\
  \midrule
\tf{QQP} & \tf{Non-paraphrase} \\
& Deletion  & 85.4 & 86.8 & 86.9 & 86.9\\
 & Addition  & 86.8 & 85.1 & 87.8 & 85.5\\
 & Change question word & 87.4 & 81.2 & 89.6 & 78.8 \\
  \cmidrule{2-6}
  & \tf{Paraphrase} \\
 & How-to questions & 90.3 & 71.2 & 92.3 & 73.0\\
& Discussion topics  & 98.4 & 56.9 & 96.2 & 66.7\\
& Same question word & 90.3 & 75.0 & 91.1 & 76.9\\
\bottomrule
\end{tabular}
}
\caption{
\label{tab:combined_errors}
We find the validation examples in SNLI and QQP containing each shortcut and partition them into \tf{Supporting} examples (S) and \tf{Counter}-examples (C) according to whether or not they have the training majority class label, and report the accuracy of BERT and RoBERTa models on each split. (See Section~\ref{sec:model_analysis}.)
}
\end{table}

%% file: tables/snli_contrasting_example_results.tex
\begin{table}[t]
\centering
\resizebox{\linewidth}{!}{%
\begin{tabular}{p{0.8\linewidth} r r}
\toprule
 \tf{Edit}  & \tf{\# Sets} & \tf{Error} \\
\midrule
  \textbf{Hypernyms} & 389 &  21.8$_{\pm0.8}$ \\
  \shortstack[l]{A man is smoking at sunset.\\
              A {\color{magenta}\sout{man}} {\color{teal}{+person}} smoking a cigarette.} \\
  \textbf{Antonyms} &  281 & 71.1$_{\pm3.8}$ \\
  \shortstack[l]{Two black dogs splash around on the beach.\\
              The dogs are playing with a {\color{teal}{+white}} ball.} \\
  \textbf{Add adjective} & 1,470 & 45.6$_{\pm8.4}$ \\
  \shortstack[l]{A man taking photos of nature.\\
                 A {\color{teal}{+sad}} man is taking photos of a wedding.} \\
\bottomrule
\end{tabular}
}
\caption{
\label{tab:snli_contrasting_example_results}
Examples of the contrastive edits we create for three shortcut features, the number of contrast sets (each consisting of one or more perturbations of a single validation instance), and the \tf{error rate} (Section~\ref{sec:model_analysis}).
We report the average and standard deviation of BERT models trained with three random seeds.
}
\end{table}

%% file: sections/experiments.tex
\input{tables/snli_robust_training_contrast_sets}
\input{figures/saliency_map}

\section{Remedying Shortcuts}
\label{sec:remedying_shortcuts}
Once we have found shortcuts, to what extent can we can mitigate them using standard robust optimization algorithms?
We conduct a small-scale experiment, focusing the contrasting examples we created for SNLI.
We compare two methods:
Just Train Twice~\citep[JTT; ][]{liu2021just} and
DRiFt~\citep{he2019unlearn}.
Note that JTT does not assume known shortcut features while DriFT does. 
Specifically, JTT upsamples the subset of training examples that are misclassified by a weak model (we use BERT-base trained for one epoch).
DRiFT takes as input a biased model and trains a new model with a regularization term that effectively upweighs the reward for instances that the biased model misclassifies (we set the biased model to be a logistic regression classifier trained on feature vectors indicating the set of SCFG production rules that appear in each tree).
Full  details are provided in Appendix~\ref{appendix:training_details}.

\paragraph{Results}
Table~\ref{tab:snli_robust_training_contrast_sets}
shows that the robust tuning methods improve performance on all the test sets, with DRiFt achieving a lower error rate than JTT. %
However, the improvements on the Antonyms test set are less substantial, which could be explained by the fact that this shortcut has few counter-examples in the training data.
These results suggest that the best solution to addressing highly correlated shortcuts may be to collect additional training data.
\input{figures/ngram_comparison_accuracy_drop}

\section{Comparing Existing Methods for Finding Shortcuts}
\label{sec:comparison_others}

\paragraph{Saliency methods}
In Table~\ref{tab:instances}, we compare our grammar-based approach with saliency heatmaps, as a method for identifying the discriminative features in an individual example.
We show two validation examples that a BERT-based classifier fails to classify.
Following~\citet{bastings2021will}, we use the L2 norm of the Integrated Gradient score~\citep{sundararajan2017axiomatic} and highlight the five tokens with the highest score.
The heatmap highlights input words but does not provide information about how different input words are connected.

To find locally important features using the grammar, we find the maximum likelihood parse trees and list the production rules in order of ${\hat{p}(\hat{y} \mid r)/\hat{p}(y^* \mid r)}$, where $y^*$ is the true label, $\hat{y}$ is the predicted label, and $\hat{p}(y \mid r)$ is proportional to the number of training trees with label $y$ that contain production rule $r$.
The most discriminative rules include production rules associated with deleting a prepositional phrase (which appear twice as often in \textit{entailment} examples) and inserting a prepositional phrase (more common in \textit{neutral} examples).
In both cases, when we delete the prepositional phrase, the model predicts the correct label.

\paragraph{N-gram-based methods}
The grammar features can capture information that cannot be expressed with n-gram features, such as alignment and syntactic roles, but how relevant is this information for diagnosing dataset shortcuts?
We explore this question in Figure~\ref{fig:ngram_comparison_accuracy_drop} by comparing the SNLI features described in Table~\ref{tab:snli_features} to simpler n-gram features created by discarding information about syntax and alignment.
For example, the \textit{Subject hypernym} feature appears in sentence pairs like ``A man is walking/A human is walking.'' 
We compare this feature to the corresponding n-gram pair feature, which has a value of true whenever ``man'' appears in the premise and ``human'' appears in the hypothesis, and would include sentences like ``A man is feeding ducks/A man is feeding a human.''
Finally, prior work has reported that individual premise words are correlated with labels in NLI datasets~\citep{poliak2018hypothesis,gururangan2018annotation}, so we also compare these features with the hypothesis-only feature, which is true whenever ``human'' appears in the second sentence.
Empty alignments, e.g. $\epsilon/w^b$, are less straightforward to compare; we define the equivalent n-gram pair feature as $w^b \in x^b$.

Figure~\ref{fig:ngram_comparison_accuracy_drop} shows that,
as we consider increasingly simple features, we identify more examples that contain the shortcut, but the shortcut becomes less discriminative, and has a weaker correlation with the BERT classifier's accuracy: BERT performs relatively worse on the supporting examples, and better on the counter-examples, indicating that these features may be less useful for diagnosing classifier errors.
More details are in Appendix~\ref{appendix:simple_features}.

%% file: tables/snli_robust_training_contrast_sets.tex
\begin{table}
\centering
\resizebox{0.90\columnwidth}{!}{%
\begin{tabular}{l c c c}
\toprule
  \tf{Method} & \tf{Hypernyms} & \tf{Antonyms} & \tf{Add adj.} \\
\midrule
BERT & 21.8$_{\pm0.8}$ & 71.1$_{\pm3.8}$ & 45.6$_{\pm8.4}$ \\
\midrule
JTT & 21.5$_{\pm0.8}$ & 69.7$_{\pm3.4}$ & 39.3$_{\pm7.8}$ \\
DRiFt & \tf{13.0}$_{\pm4.1}$ & \tf{68.7}$_{\pm6.8}$ & \tf{29.4}$_{\pm4.5}$ \\
\bottomrule
\end{tabular}
}
\caption{
\label{tab:snli_robust_training_contrast_sets}
Evaluating robust training methods on the test sets described in Table~\ref{tab:snli_contrasting_example_results}.
We report the mean error rate ($\downarrow$) and standard deviation from three runs (Section~\ref{sec:remedying_shortcuts}).
}
\end{table}

%% file: figures/saliency_map.tex
\begin{table*}[ht]
\centering
\resizebox{1.0\linewidth}{!}{%
\begin{tabular}{m{0.4\linewidth} l l l} %
\toprule
{\Large \tf{Saliency map}} & {\Large \tf{Tree parse}}
& {\Large \tf{Label}}
& {\Large \tf{Prediction}} \\
\midrule
  
{
\large
\sethlcolor{saliencycolor}
    A woman {\hl{pushing}} a {\hl{coffee}} cart through a plaza.
    A {\hl{coffee}} {\hl{worker}} {\hl{pushes}} a cart.
}
    &
    \begin{tabular}{l}
    \tikzset{delink/.style={
        decorate,
        postaction={decorate,
        decoration={markings, mark=at position 0.5 with {
        \draw[-] ++ (-3pt,-3pt) -- (3pt,3pt);}
        }
        }}}
      \begin{forest} 
        for tree={
            fit=tight,
            calign=fixed edge angles,calign angle=75,
            l sep-=1.5em,
            font=\large,
            if n children=0{
              rectangle,
              rounded corners=5pt,
              fill=patterncolor,
              tier=terminal,
              align=center,
              l=2em
            }{
            circle,
            draw,
            inner sep=0,
            minimum height=12pt,
            l=1em
            },
        }
        [{1}
          [{14}, 
            [A woman/ \\A coffee worker, roof]]
          [{30},highlight,tikz={\node [draw, dotted, thick, rounded corners=5pt, inner xsep=2pt, edgehighlightcolor, fit=()(!11)(!ll)]{};},
              [{21}
                [pushing a coffee cart/ \\ pushes a cart, roof]]
              [{10}, [through a plaza/ \\ $\epsilon$, roof]]
          ]]
    \end{forest}
\end{tabular}
& {\large \textexneut{Neutral}} & {\large \textexpos{Entailment}} \\
{
\large
\sethlcolor{saliencycolor}
    A man kneels {\hl{next}} to a colorful display {\hl{outside}}.
    The man is planting a backyard {\hl{garden}} during {\hl{spring}} {\hl{.}}
}
    &
    \begin{tabular}{l}
    \tikzset{delink/.style={
        decorate,
        postaction={decorate,
        decoration={markings, mark=at position 0.5 with {
        \draw[-] ++ (-3pt,-3pt) -- (3pt,3pt);}
        }
        }}}
      \begin{forest} 
        for tree={
            fit=tight,
            calign=fixed edge angles,calign angle=75,
            l sep-=1.5em,
            font=\large,
            if n children=0{
              rectangle,
              rounded corners=5pt,
              fill=patterncolor,
              tier=terminal,
              align=center,
              l=2em
            }{
            circle,
            draw,
            inner sep=0,
            minimum height=12pt,
            l=1em
            },
        }
        [{1}
          [{1},
            [A man kneels next to a colorful display outside/ \\ The man is planting a backyard garden, roof]]
          [{17},highlight,tikz={\node [draw, thick, dotted, rounded corners=5pt, inner xsep=2pt, edgehighlightcolor, fit=()(!11)(!ll)]{};},
            [{49} [{$\epsilon$/\\during}] ]
            [{30} [{$\epsilon$/\\spring}] ]]]
    \end{forest}
\end{tabular}
& {\large \textexneg{Contradiction}} & {\large \textexneut{Neutral}} \\
    \bottomrule
\end{tabular}
}

\caption{Saliency maps and SCFG parse trees for two SNLI validation instances that are misclassified by a fine-tuned BERT model (Section~\ref{sec:comparison_others}).
For each tree, we highlight the production rule most strongly correlated with the predicted label, which is associated with the pattern of either deleting or inserting a prepositional phrase.
In both cases, when we delete the prepositional phrase, the model predicts the correct label.
}
\label{tab:instances}
\end{table*}

%% file: figures/ngram_comparison_accuracy_drop.tex
\begin{figure*}[ht]
  \centering
  \includegraphics[width=0.97\linewidth]{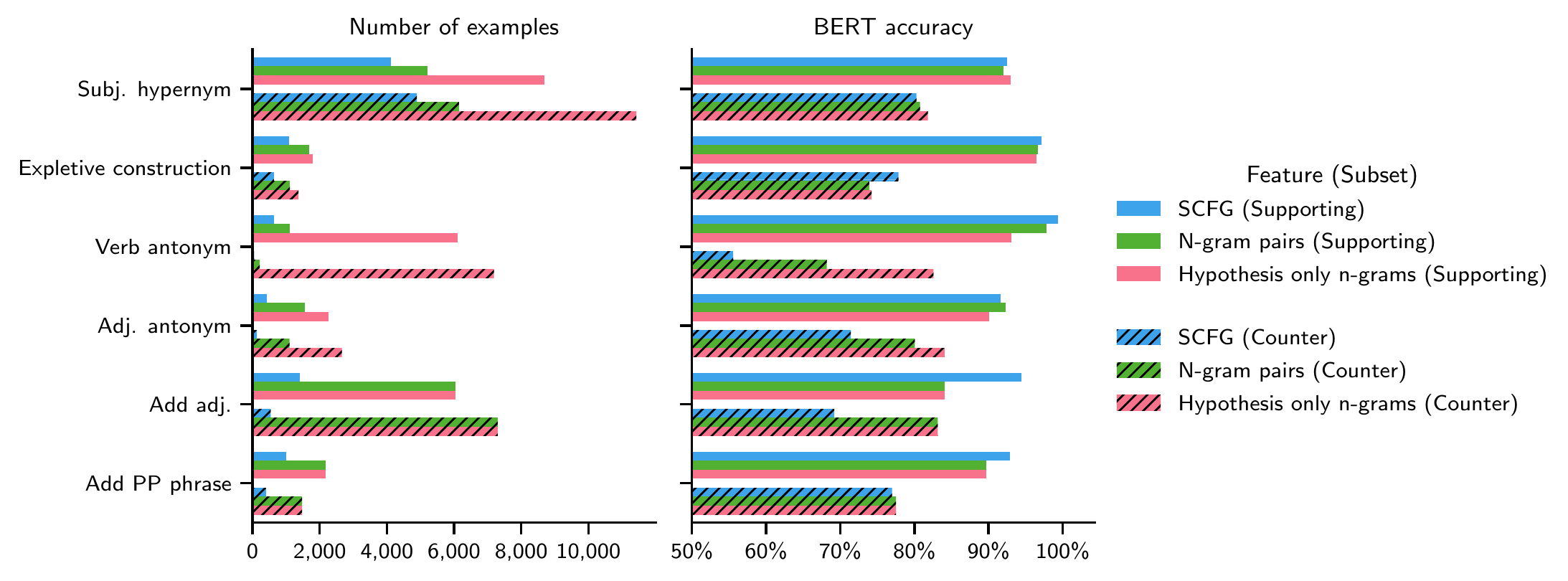}
\caption{
We compare the SCFG features from SNLI with equivalent n-gram feature that discard the alignment information provided by the grammar (Section~\ref{sec:comparison_others}).
As we consider simpler features, the features appear in more examples but become less discriminative, and tend to have a weaker correlation with BERT accuracy.
}
\label{fig:ngram_comparison_accuracy_drop}
\end{figure*}

%% file: sections/related.tex
\section{Related Work}
\label{sec:related}

\vspace{-1em}
\paragraph{Finding shortcuts}
Prior work has identified shortcuts in NLP datasets by developing diagnostic evaluation datasets~\citep{mccoy2019right,niven2019probing,rosenman2020exposing}, training partial input baselines~\citep{gururangan2018annotation,poliak2018hypothesis},
or calculating statistics of simple features, like n-grams, that can be enumerated explicitly~\citep[e.g.][]{schuster2019towards,gardner2021competency}.
~\citet{han2020explaining} use influence functions~\citep{koh2017understanding} to retrieve training instances that are relevant to a model's prediction. \citet{pezeshkpour2022combining} combine instance attribution with feature attribution, which highlights salient input tokens.
In contrast, our aim is to find dataset-level shortcuts that can be expressed as explicit feature functions. %

\paragraph{Defining spurious correlations}
A related line of work has addressed the question of which correlations should be considered spurious.
~\citet{gardner2021competency} suggest that all correlations between labels and low-level features, such as unigrams, are spurious.
~\citet{eisenstein2022informativeness} argues that such correlations will arise naturally in most language classification settings, and domain expertise is needed to determine which correlations might be harmful.
Our aim is to provide a summary of the shortcuts in a dataset, so that a practitioner can determine which are undesirable and remedy them if needed.

\paragraph{Feature importance}
Our approach is related to methods for identifying important input features, like LIME~\citep{ribeiro2016should} .
~\citet{lundberg2017unified} present a framework for measuring local feature importance based on Shapley values~\citep{shapley1953value}, which can be applied to both model-specific and model-agnostic notions of importance, and~\citet{covert2020understanding} extend this approach to global importance scores.
These methods require first identifying the set of features,
while our focus is on inducing a richer set of features.
We use a simple metric for feature importance, and leave more sophisticated metrics to future work.

\paragraph{Grammar induction}
Grammar induction has been a long-standing subject of research in artificial intelligence. Prior work has used grammar induction for linguistic
analysis~\citep{johnson2008unsupervised,dunn2018finding} and feature
extraction~\citep{hardisty2010modeling,wong2012exploring}.
Synchronous and quasi-synchronous grammars~\citep{smith2006quasi} have been used in machine translation and applied to a variety of other NLP tasks~\citep[e.g.][]{wang2007jeopardy,blunsom2008bayesian,yamangil2010bayesian}, but have largely been supplanted by end-to-end neural network approaches.
 \citet{kim2021sequence} develop a neural quasi-synchronous grammar as an interpretable, rule-based model for sequence transduction tasks.
Our work shares a similar motivation, but applied to a different goal, modeling dataset shortcuts.

%% file: sections/appendix.tex
\clearpage

\section{Grammar Parameterization}
\label{appendix:parameterization}
Let $\gN, \gV, \gR$ denote the set of non-terminal symbols, terminal symbols, and production rules in the grammar.
We only consider grammars in binary normal form.
Following~\citet{kim2019compound}, we designate a subset $\gP \subseteq \gN$ as pre-terminals.
$R$ is then defined as all rules of the form $\alpha \to w$, where $\alpha \in \gP$ and $w \in \gV$, or $\alpha \to \beta \; \gamma$, where $\alpha \in \gN \setminus \gP$ and $\beta, \gamma \in \gN$, as well as $s \to \alpha$, where $\alpha \in \gN$ and $S$ is the start symbol.

We parameterize our grammars using neural networks, following~\citet{kim2019compound} and~\citet{kim2021sequence}.
For the PCFG, we use the parameterization from the neural PCFG of~\citet{kim2019compound}:
\begin{align*}
  & p(S \to \alpha) &\propto& \exp(f_1(\vv_S)^{\top}\vu_{S\to \alpha})\\
  & p(\alpha \to \beta \; \gamma) &\propto& \exp(f_2(\vv_{\alpha})^{\top}\vu_{\beta\gamma})\\
  & p(\alpha \to w) &\propto& \exp(f_3(\vv_{\alpha})^{\top}\vu_{w}),\\
\end{align*}
where $\vv_*, \vu_* \in \R^d$ are embeddings and $f_1, f_2, f_3$ are multilayer perceptrons with two hidden layers and residual connections.

For the SPCFG, we parameterize the the starting rules $S \to \alpha$ and the binary productions $\alpha \to \beta \; \gamma$ the same as in the PCFG.
We parameterize the terminal productions $\alpha \to w$ by factoring according to the kind of emission, $k \in \{w^a/w^b, w^a/\epsilon, \epsilon/w^b, \text{copy}\}$, where $\epsilon$ is the empty symbol and $\text{copy}$ is a production with $w^a/w^b$ with $w^a = w^b$. The emission distribution is then factored as:
\begin{align*}
  p(\alpha \to w^a/w^b) = &\;p(k \mid \alpha) \times p(w^a \mid \alpha, k) \\
                         &\times p(w^b \mid \alpha, k, w^a).
\end{align*}
We parameterize these rules as in the PCFG:
\begin{align*}
   &p(k \mid \alpha) &\propto& \exp(f_3(\vv_{\alpha^k})^{\top}\vu_k)\\
   &p(w^a \mid k, \alpha) &\propto& \exp(f_4(\vv_{\alpha^a})^{\top}\vu_{w^a})\\
   &p(w^b \mid k, \alpha, w^a) &\propto& \exp(f_5(\vv_{\alpha^b} + \vv_{w^a})^{\top}\vu_{w^b}),
\end{align*}
with restrictions to ensure that the distribution is of the emission kind $k$.

\section{Training Details}
\label{appendix:training_details}

\paragraph{Grammar induction}
For all of our experiments, we fix the size of the grammar at 64 pre-terminal symbols and 32 non-terminal symbols. We set the hidden dimension $d$ of the grammar embeddings to be 256 and define every $f_i$ to be a ReLU network with two hidden layers, following~\citet{kim2019compound}.
We use a learning rate of 1e-3 and the Adam optimizer.
For the PCFGs, we train for up to 40 epochs, evaluating on validation data every 4096 steps and stopping early after five checkpoints with no improvement in validation loss (negative log likelihood).
For the SCFGs, we train for up to 10 epochs and use the same early stopping policy.

We run our main experiments on four RTX 3090 Ti GPUs with 24GB of memory each, but also test our SCFG implementation on RTX 2080 GPUs with 12GB of memory.
With four GPUs, and training on 64,000 examples with $|x^a| \times |x^b| \leq 225$, running SCFG grammar induction until convergence and then parsing all of the training and validation examples takes between 12 and 24 hours hours.
For the PCFGs, we use a mini-batch size of four sequences per GPU, and for the SCFG we use a mini-batch size of one pair of sequences per GPU.
Consistent with the report of~\citet{kim2021sequence}, we find that GPU memory is the main bottleneck to scaling the SCFG.

\paragraph{Classification experiments}
We train BERT and RoBERTa classifiers using a learning rate of 1e-5 and do not tune any hyperparameters.
The biased model for DRiFt is a logistic regression model trained with L1 regularization using the default implementation in scikit-learn~\cite{pedregosa2011scikit}, and the identification model for Just-Train-Twice is a BERT model trained for one epoch.
We use a mini-batch size of 16 and training these models for up to 20 epochs, evaluating on the validation set every 4096 steps, stopping early after five checkpoints with no improvement in validation accuracy.

\section{Comparing Simple Features}
\label{appendix:simple_features}

In Table~\ref{tab:app_group_feature_statistics_snli}, we compare our SCFG features to corresponding n-gram features in terms of prevalence, accuracy, and the correlation with BERT's error rate on SNLI and QQP. The aim of this experiment is to estimate the extent to which the BERT classifiers exploit the type of syntactic patterns identified by the grammar or, for example, discard positional information.

\input{tables/app_group_feature_statistics_snli}

\section{Additional Analysis}
\label{appendix:additional_results}

\subsection{Additional Features}
\label{appendix:additional_features}
\input{tables/app_imdb_additional_features}
\input{tables/app_subj_additional_features}
\input{tables/app_snli_additional_features}
\input{tables/app_qqp_additional_features}
We list additional features for IMDb (Table~\ref{tab:additional_imdb_features}), SUBJ (Table~\ref{tab:additional_subj_features}), SNLI (Table~\ref{tab:additional_snli_features}), and QQP (Table~\ref{tab:additional_qqp_features}).
For each table, we pick the 1,000 subtrees with the highest mutual information scores (restricted to subtrees with two or more leaves) and group them by root label $\alpha$ and majority class label $y$.
For each subtree $s$, the corresponding feature $Z_s$ is the boolean feature defined as 1 if subtree $s$ appears in the maximum likelihood parse tree and 0 otherwise, and we calculate the mutual information using the empirical likelihoods, $\hat{p}(y, s) \propto 1 + \sum_{t_i, y_i} \mathbbm{1}[y_i = y \wedge s \in t_i]$, where $t_i$ represents the parse for input $x_i$.
Each row represents a composite feature $Z_{\alpha, y} = \bigvee_{s \in S(\alpha, y)} Z_s$, where $S$ is the subset of the top 1,000 subtrees that have root label $\alpha$ and majority class label $y$.
We list the rows in decreasing order of $I(Z_{\alpha, y}; Y)$.
For each row, we report the number of training examples with $Z_{\alpha, y} = 1$ for each class label, and list the spans corresponding to up to five subtrees $s \in S(\alpha, y)$, in decreasing order of $I(Z_s; Y)$.

\subsection{Higher Level Features}
\input{tables/snli_nonterminals}

Table~\ref{tab:snli_nonterminals} lists the twelve binary production rules $\alpha \to \beta \; \gamma$ that have the highest mutual information in SNLI.
For each rule $r$, the corresponding feature $Z_r$ is the boolean feature defined as 1 if production rule $r$ appears in the maximum likelihood parse tree and 0 otherwise, and we calculate the mutual information using the empirical likelihoods, $\hat{p}(y, r) \propto 1 + \sum_{t_i, y_i} \mathbbm{1}[y_i = y \wedge r \in t_i]$, where $t_i$ represents the parse for input $x_i$.
The most informative features include removing a prepositional phrase (entailment) and adding an object or prepositional phrase (neutral).

\subsection{Contrast Sets}
\label{appendix:additional_contrast sets}
\input{tables/snli_contrasting_example}

In addition to creating rule-based contrast sets, we also create a set of contrasting examples manually according to the following procedure, illustrated in Table~\ref{tab:snli_contrasting_example}. For a given shortcut feature $Z$ associated with majority label $y^{Z=1}$, we start by identifying validation examples $(x_0, y_0)$ for which $Z = 1$ and $y_0 = y^{Z=1}$, and simplifying $x_0$ by removing other features that support $y^{Z=1}$, getting a new instance $x_1$ with $y_1 = y^{Z=1}$, for which $Z = 1$ provides the most evidence in favor of the label. Then we make a series of small changes to get examples $x', y'$ with $y' \neq y_z$: first, several control examples with $Z \neq 0$ to verify that our perturbations change the model's behavior as expected in the absence of the feature ($x_{2, \ldots, 4}$), and finally an instance with $Z = 1$ ($x_5$).

We apply this procedure to the ``Adjective antonym'' feature, illustrated in Table~\ref{tab:snli_contrasting_example}.
This feature is associated with the pre-terminal label \rootlabel{85} and includes productions like \textex{white/black} and \textex{red/blue}.
Our test is to modify the premise by moving the adjective from the subject noun to the object noun, which changes the label from \ti{contradiction} to \ti{neutral}.
We randomly select ten validation examples that contain the feature and have the majority label, contradiction. For each example, we create four control examples and one test example, as in Table~\ref{tab:snli_contrasting_example}.
The BERT model makes the expected prediction for 40/40 control examples but misclassifies 9/10 test examples, in each case predicting contradiction rather than neutral.
This indicates that the model predicts contradiction when the premise and hypothesis contain contradicting adjectives, even if the adjectives describe different entities.

%% file: tables/app_group_feature_statistics_snli.tex
\begin{table*}[ht]
\centering
\resizebox{\linewidth}{!}{%
\small
\begin{tabular}{l l rrrrrr | rrrrrr}
\toprule
&  & \multicolumn{6}{c|}{\textbf{\ti{Number of examples}}} & \multicolumn{6}{c}{\textbf{\ti{BERT accuracy}}} \\
&  & \multicolumn{2}{c}{\textbf{SCFG}} & \multicolumn{2}{c}{\textbf{N-gram pairs}} & \multicolumn{2}{c|}{\textbf{Hyp. only}} & \multicolumn{2}{c}{\textbf{SCFG}} & \multicolumn{2}{c}{\textbf{N-gram pairs}} & \multicolumn{2}{c}{\textbf{Hyp. only}} \\
&  & S & C & S & C & S & C & S & C & S & C & S & C \\
\midrule
\tf{SNLI} & \textbf{Entailment} \\
& Subj. phrase hypernym & 4,115 & 4,894 & 5,204 & 6,149 & 8,698 & 11,427 & 92.5 & 80.3 & 92.0 & 80.8 & 93.0 & 81.8 \\
& Copy verb & 908 & 418 & 3,151 & 2,990 & 6,224 & 9,884 & 98.6 & 83.6 & 95.5 & 81.5 & 92.7 & 82.0 \\
& Expletive construction & 1,087 & 638 & 1,682 & 1,112 & 1,782 & 1,368 & 97.2 & 77.8 & 96.7 & 73.9 & 96.4 & 74.2 \\
\cmidrule{2-14}
& \textbf{Contradiction} \\
& Subj. antonym & 1,127 & 108 & 2,178 & 1,121 & 10,243 & 18,129 & 96.9 & 61.5 & 95.5 & 81.1 & 88.2 & 84.2 \\
& Subj. phrase antonym & 1,116 & 235 & 1,784 & 1,148 & 9,156 & 15,327 & 98.2 & 77.8 & 94.0 & 83.8 & 88.8 & 84.7 \\
& Verb antonym & 645 & 50 & 1,107 & 215 & 6,104 & 7,188 & 99.4 & 55.6 & 97.8 & 68.2 & 93.1 & 82.6 \\
& Definite article & 6,055 & 9,381 & 10,803 & 17,470 & 10,803 & 17,470 & 86.2 & 82.2 & 80.3 & 85.7 & 80.3 & 85.7 \\
& Adj. antonym & 432 & 128 & 1,565 & 1,099 & 2,262 & 2,658 & 91.7 & 71.4 & 92.3 & 80.1 & 90.1 & 84.1 \\
\cmidrule{2-14}
& \textbf{Neutral} \\
& Add function word & 7,317 & 7,161 &  &  & 21,024 & 40,092 & 86.6 & 81.0 &  &  & 78.9 & 87.7 \\
& Add object noun & 2,964 & 1,711 &  &  & 16,343 & 29,792 & 89.0 & 76.1 &  &  & 80.1 & 86.9 \\
& Add adjective & 1,404 & 541 &  &  & 6,044 & 7,298 & 94.4 & 69.2 &  &  & 84.1 & 83.1 \\
& Add PP phrase & 1,009 & 402 &  &  & 2,186 & 1,473 & 92.9 & 77.0 &  &  & 89.7 & 77.5 \\
\midrule
\tf{QQP} & \tf{Paraphrase} \\
& Discussion topic & 1,806 & 373 & 2,210 & 480 & 2,891 & 963 & 98.4 & 56.9 & 98.3 & 58.1 & 98.2 & 68.3 \\
& How-to question & 6,415 & 3,269 & 6,964 & 3,587 & 17,123 & 14,370 & 90.3 & 71.2 & 90.5 & 71.7 & 90.6 & 78.5 \\
& Same question word & 10,403 & 6,861 & 11,485 & 7,634 & 25,302 & 24,279 & 90.3 & 75.0 & 90.3 & 75.1 & 89.5 & 80.6 \\
\bottomrule
\end{tabular}
}
\caption{
\label{tab:app_group_feature_statistics_snli}
We compare the SCFG features from SNLI (top) and QQP (bottom) with an equivalent pair-of-n-gram feature that discards the alignment information provided by the grammar (see Section~\ref{sec:comparison_others}).
For example, let $x^a, x^b$ denote the first and second sentence in a pair, and $t$ denote the maximum likelihood SCFG tree.
In the first row, the \tf{SCFG} feature represents the indicator $\mathbbm{1}[(\rootlabel{14} \; \text{a man/a person}) \in t \vee (\rootlabel{14} \; \text{a woman/a human}) \in t \vee \ldots]$; the \tf{N-gram pair feature} represents the indicator $\mathbbm{1}[(\text{a man} \in x^a \wedge \text{a person} \in x^b) \vee (\text{a woman} \in x^a \wedge \text{a human} \in x^b) \vee \ldots]$; and the Hypothesis-only n-gram feature (\tf{Hyp. only}) represents  $\mathbbm{1}[(\text{a person} \in x^b) \vee (\text{a human} \in x^b) \vee \ldots]$.
In the case of an empty alignment, e.g. $w^a/\epsilon$, the equivalent n-gram feature is defined as $w^a \in x^a$.
For each feature $Z$, we find the examples for which $Z=1$ and partition them into supporting examples (\tf{S}) and counter-examples (\tf{C}) according to whether or not they have the class label $y$ that appears most often in the training example for which $Z=1$.
We report the number of training examples in each subset and the accuracy of a BERT classifier on corresponding validation examples.
The simpler features appear in more examples but tend to be less discriminative and to have a weaker correlation with the BERT classifier's accuracy: BERT performs relatively worse on the supporting examples, and better on the counter-examples, indicating that these features may be less useful for diagnosing classifier errors.
In QQP, the grammar features do not convey much more information than n-gram pairs, perhaps indicating that syntactic alignment is relatively unimportant for identifying paraphrases in this dataset.
}
\end{table*}

%% file: tables/app_imdb_additional_features.tex
\begin{table*}
\centering
\resizebox{\linewidth}{!}{%
\begin{tabular}{lp{\linewidth}rr}
\toprule
\tf{Root} & \tf{Examples} & \textexneg{\tf{N}} & \textexpos{\tf{P}} \\
\midrule
\rootlabel{25} & \textexneg{so bad}, \textexneg{waste your time}, \textexneg{not funny}, \textexneg{even worse}, \textexneg{the worst movie} & 5,448 & 1,562 \\
\rootlabel{25} & \textexpos{highly recommended}, \textexpos{a must see}, \textexpos{a great movie}, \textexpos{a great job}, \textexpos{a great film} & 1,484 & 4,176 \\
\rootlabel{10} & \textexneg{waste of time}, \textexneg{bad movie}, \textexneg{good thing}, \textexneg{terrible movie}, \textexneg{horror movie} & 1,904 & 297 \\
\rootlabel{10} & \textexpos{must see}, \textexpos{great job}, \textexpos{great movie}, \textexpos{great film}, \textexpos{wonderful movie} & 678 & 2,530 \\
\rootlabel{18} & \textexneg{at all}, \textexneg{at all costs}, \textexneg{at least}, \textexneg{at best}, \textexneg{than this} & 5,272 & 2,710 \\
\rootlabel{31} & \textexneg{don ' t waste your time}, \textexneg{i mean}, \textexneg{don ' t bother}, \textexneg{it fails}, \textexneg{it was so bad} & 2,107 & 463 \\
\rootlabel{21} & \textexneg{worst movie}, \textexneg{worst film}, \textexneg{worst movies}, \textexneg{piece of crap}, \textexneg{worst films} & 2,368 & 646 \\
\rootlabel{31} & \textexpos{i loved it}, \textexpos{i recommend it}, \textexpos{i love this movie}, \textexpos{i loved this movie}, \textexpos{i highly recommend it} & 191 & 1,377 \\
\rootlabel{18} & \textexpos{on dvd}, \textexpos{as well}, \textexpos{in love}, \textexpos{at the same time}, \textexpos{for everyone} & 1,471 & 3,337 \\
\rootlabel{15} & \textexneg{your time}, \textexneg{your money}, \textexneg{all costs}, \textexneg{the worst movies}, \textexneg{this crap} & 7,836 & 5,567 \\
\rootlabel{15} & \textexpos{the same time}, \textexpos{all ages}, \textexpos{the best movies}, \textexpos{the best}, \textexpos{the show} & 2,413 & 4,383 \\
\rootlabel{30} & \textexpos{well -}, \textexpos{must -}, \textexpos{heart -}, \textexpos{fun ,}, \textexpos{fun and} & 647 & 2,039 \\
\rootlabel{30} & \textexneg{really bad}, \textexneg{boring ,}, \textexneg{dull ,}, \textexneg{low budget}, \textexneg{so -} & 2,700 & 1,069 \\
\rootlabel{16} & \textexneg{3 /}, \textexneg{4 /}, \textexneg{2 /}, \textexneg{1 /}, \textexneg{1 out of} & 1,673 & 450 \\
\rootlabel{9} & \textexneg{don ' t}, \textexneg{i ' m}, \textexneg{there was}, \textexneg{the acting is}, \textexneg{i could} & 9,203 & 7,415 \\
\rootlabel{7} & \textexpos{loved it}, \textexpos{love this movie}, \textexpos{loved this movie}, \textexpos{recommend it}, \textexpos{enjoyed it} & 906 & 2,229 \\
\rootlabel{24} & \textexneg{of time}, \textexneg{of crap}, \textexneg{of the worst movies}, \textexneg{of my life}, \textexneg{of the worst films} & 3,306 & 1,692 \\
\rootlabel{17} & \textexneg{at all .}, \textexneg{at all costs .}, \textexneg{instead .}, \textexneg{whatsoever .}, \textexneg{? ? ?} & 2,333 & 976 \\
\rootlabel{29} & \textexpos{10 / 10}, \textexpos{8 / 10}, \textexpos{7 / 10}, \textexpos{highly recommended .}, \textexpos{9 / 10} & 5 & 481 \\
\rootlabel{7} & \textexneg{be funny}, \textexneg{work with}, \textexneg{sit through}, \textexneg{be a comedy}, \textexneg{waste your time} & 1,525 & 468 \\
\rootlabel{19} & \textexneg{the acting}, \textexneg{this movie}, \textexneg{the plot}, \textexneg{the script}, \textexneg{it just} & 5,849 & 3,993 \\
\rootlabel{16} & \textexpos{10 /}, \textexpos{8 /}, \textexpos{7 /}, \textexpos{9 /}, \textexpos{7 out of} & 396 & 1,304 \\
\rootlabel{13} & \textexneg{bad acting}, \textexneg{bad movies}, \textexneg{special effects}, \textexneg{poor acting}, \textexneg{terrible acting} & 1,623 & 572 \\
\rootlabel{0} & \textexneg{don '}, \textexneg{couldn '}, \textexneg{didn '}, \textexneg{wasn '}, \textexneg{can '} & 6,426 & 4,611 \\
\rootlabel{5} & \textexpos{walter matthau}, \textexpos{james stewart}, \textexpos{jon voight}, \textexpos{william powell}, \textexpos{philo vance} & 85 & 666 \\
\rootlabel{29} & \textexneg{4 / 10}, \textexneg{3 / 10}, \textexneg{1 / 10}, \textexneg{2 / 10}, \textexneg{1 / 2 from * * * *} & 416 & 13 \\
\rootlabel{8} & \textexneg{30 minutes}, \textexneg{five minutes}, \textexneg{10 minutes}, \textexneg{90 minutes}, \textexneg{2 hours} & 1,083 & 329 \\
\rootlabel{21} & \textexpos{same time}, \textexpos{best movies}, \textexpos{first time}, \textexpos{best movie}, \textexpos{best film} & 288 & 966 \\
\rootlabel{24} & \textexpos{of life}, \textexpos{of the best}, \textexpos{of the best movies}, \textexpos{of fun}, \textexpos{of my favorites} & 579 & 1,424 \\
\rootlabel{9} & \textexpos{it is}, \textexpos{i highly}, \textexpos{i first}, \textexpos{you will}, \textexpos{this is} & 5,595 & 6,748 \\
\rootlabel{1} & \textexpos{my favorite}, \textexpos{his best}, \textexpos{today ' s}, \textexpos{my only}, \textexpos{my all time} & 655 & 1,386 \\
\rootlabel{17} & \textexpos{together .}, \textexpos{as well .}, \textexpos{today .}, \textexpos{very well .}, \textexpos{too .} & 392 & 978 \\
\rootlabel{28} & \textexneg{' m}, \textexneg{' re}, \textexneg{' t}, \textexneg{' d}, \textexneg{` s} & 3,283 & 2,200 \\
\rootlabel{22} & \textexneg{avoid this movie}, \textexneg{first of all}, \textexneg{i have ever seen}, \textexneg{save your money}, \textexneg{skip this one} & 689 & 248 \\
\rootlabel{13} & \textexpos{great performances}, \textexpos{great acting}, \textexpos{excellent performances}, \textexpos{twists and turns}, \textexpos{great fun} & 100 & 410 \\
\rootlabel{14} & \textexpos{and enjoy}, \textexpos{and sad}, \textexpos{10 / 10}, \textexpos{, as always}, \textexpos{worth watching} & 47 & 277 \\
\rootlabel{11} & \textexneg{" film}, \textexneg{" movie}, \textexneg{" plot}, \textexneg{" comedy}, \textexneg{" so bad it ' s good} & 207 & 19 \\
\rootlabel{5} & \textexneg{ed wood}, \textexneg{steven seagal}, \textexneg{van damme}, \textexneg{uwe boll}, \textexneg{tom savini} & 167 & 7 \\
\rootlabel{14} & \textexneg{or something}, \textexneg{. . .}, \textexneg{and boring}, \textexneg{, right}, \textexneg{and pointless} & 989 & 501 \\
\rootlabel{6} & \textexneg{. .}, \textexneg{. . .}, \textexneg{. . well}, \textexneg{. . no}, \textexneg{. . oh} & 2,753 & 1,935 \\
\rootlabel{19} & \textexpos{this game}, \textexpos{the series}, \textexpos{my only complaint}, \textexpos{the film}, \textexpos{it also} & 1,435 & 2,024 \\
\rootlabel{1} & \textexneg{their right}, \textexneg{your time or}, \textexneg{someone ' s}, \textexneg{your time and}, \textexneg{your local} & 174 & 31 \\
\rootlabel{23} & \textexneg{it off}, \textexneg{me wrong}, \textexneg{it up}, \textexneg{through the whole thing}, \textexneg{down the toilet} & 465 & 205 \\
\rootlabel{22} & \textexpos{a must see}, \textexpos{highly recommended}, \textexpos{as always}, \textexpos{i think}, \textexpos{of course} & 504 & 793 \\
\rootlabel{23} & \textexpos{out on dvd}, \textexpos{me away}, \textexpos{- on} & 12 & 81 \\
\rootlabel{20} & \textexpos{' n} & 6 & 29 \\
\rootlabel{8} & \textexpos{many years} & 29 & 66 \\
\rootlabel{0} & \textexpos{you don '} & 212 & 282 \\
\bottomrule
\end{tabular}
}
\caption{\label{tab:additional_imdb_features}
Additional IMDb features (see Section~\ref{appendix:additional_features}).
We report the number of \ti{positive} (\tf{P}) and \ti{negative} (\tf{N}) training examples associated with each feature and highlight the features according to the most common class.
Many features are related to clear sentiment markers like adjectives, but it is also easy to identify features corresponding to numerical ratings and other patterns, like actor names, that we might not expect to be correlated with class labels.
}
\end{table*}

%% file: tables/app_subj_additional_features.tex
\begin{table*}
\centering
\resizebox{\linewidth}{!}{%
\begin{tabular}{lp{\linewidth}rr}
\toprule
\tf{Root} & \tf{Examples} & \textexneg{\tf{S}} & \textexpos{\tf{O}} \\
\midrule
\rootlabel{27} & \textexpos{his life}, \textexpos{his wife}, \textexpos{his father}, \textexpos{his mother}, \textexpos{their lives} & 323 & 1,305 \\
\rootlabel{27} & \textexneg{a movie}, \textexneg{the film}, \textexneg{the movie}, \textexneg{this movie}, \textexneg{the screen} & 846 & 134 \\
\rootlabel{13} & \textexneg{the movie}, \textexneg{but it}, \textexneg{the film}, \textexneg{if you}, \textexneg{if it} & 624 & 78 \\
\rootlabel{28} & \textexpos{decides to}, \textexpos{order to}, \textexpos{" "}, \textexpos{has been}, \textexpos{begins to} & 133 & 614 \\
\rootlabel{2} & \textexpos{" "}, \textexpos{best friend}, \textexpos{young man}, \textexpos{young [UNK]}, \textexpos{[UNK] girl} & 48 & 423 \\
\rootlabel{3} & \textexpos{finds himself}, \textexpos{finds out}, \textexpos{falls in love}, \textexpos{is [UNK]}, \textexpos{is sent} & 57 & 403 \\
\rootlabel{28} & \textexneg{' t}, \textexneg{' s}, \textexneg{' re}, \textexneg{' ll}, \textexneg{' s not} & 1,396 & 753 \\
\rootlabel{2} & \textexneg{[UNK] movie}, \textexneg{[UNK] film}, \textexneg{running time}, \textexneg{romantic comedy}, \textexneg{[UNK] plot} & 279 & 14 \\
\rootlabel{13} & \textexpos{the two}, \textexpos{when he}, \textexpos{where he}, \textexpos{the gang}, \textexpos{the girls} & 16 & 234 \\
\rootlabel{1} & \textexneg{. .}, \textexneg{as [UNK]}, \textexneg{, too}, \textexneg{in a way}, \textexneg{in the right place} & 359 & 85 \\
\rootlabel{8} & \textexpos{ [UNK] [UNK]}, \textexpos{ [UNK] [UNK] }, \textexpos{ [UNK] [UNK] }, \textexpos{[UNK] [UNK] [UNK]}, \textexpos{' s mother} & 22 & 204 \\
\rootlabel{1} & \textexpos{in love}, \textexpos{" "}, \textexpos{with him}, \textexpos{with her}, \textexpos{for her} & 61 & 291 \\
\rootlabel{3} & \textexneg{comes off}, \textexneg{' s hard}, \textexneg{makes up}, \textexneg{' d expect}, \textexneg{doesn ' t} & 176 & 29 \\
\rootlabel{31} & \textexneg{. . .}, \textexneg{. '}, \textexneg{of life .}, \textexneg{in its [UNK] .}, \textexneg{in years .} & 203 & 50 \\
\rootlabel{6} & \textexneg{' s ]}, \textexneg{' ve seen}, \textexneg{' s also}, \textexneg{' t [UNK]}, \textexneg{' t seen} & 160 & 34 \\
\rootlabel{8} & \textexneg{- -}, \textexneg{' s film}, \textexneg{and [UNK]}, \textexneg{or [UNK]}, \textexneg{- [UNK] [UNK]} & 428 & 209 \\
\rootlabel{12} & \textexneg{of the film}, \textexneg{of a movie}, \textexneg{of the year}, \textexneg{of a [UNK]}, \textexneg{of the plot} & 82 & 5 \\
\rootlabel{12} & \textexpos{of his father}, \textexpos{of their own}, \textexpos{of his life}, \textexpos{of the world}, \textexpos{of the [UNK]} & 31 & 128 \\
\rootlabel{25} & \textexpos{one day}, \textexpos{he is [UNK]}, \textexpos{along the way}, \textexpos{at the same time}, \textexpos{in the meantime} & 1 & 40 \\
\rootlabel{25} & \textexneg{it ' s [UNK]}, \textexneg{it ' s}, \textexneg{that ' s}, \textexneg{for the most part}, \textexneg{the film [UNK]} & 33 & 1 \\
\rootlabel{6} & \textexpos{" " [UNK]}, \textexpos{' s got}, \textexpos{order to [UNK]}, \textexpos{struggle to find}, \textexpos{" " tells} & 5 & 43 \\
\rootlabel{31} & \textexpos{" " .}, \textexpos{on him .}, \textexpos{of it .}, \textexpos{for [UNK] .}, \textexpos{in the [UNK] .} & 2 & 34 \\
\rootlabel{29} & \textexpos{the [UNK]}, \textexpos{her [UNK]}, \textexpos{his [UNK]}, \textexpos{their [UNK]}, \textexpos{two [UNK]} & 61 & 112 \\
\rootlabel{29} & \textexneg{its [UNK]}, \textexneg{this [UNK]} & 27 & 5 \\
\rootlabel{15} & \textexpos{sang - woo}, \textexpos{daniel [UNK] }, \textexpos{played by [UNK]} & 0 & 12 \\
\rootlabel{23} & \textexneg{[UNK] [UNK]}, \textexneg{- and} & 33 & 13 \\
\rootlabel{10} & \textexneg{- [UNK] [UNK]} & 6 & 1 \\
\rootlabel{22} & \textexneg{the most part} & 4 & 0 \\
\rootlabel{11} & \textexneg{silence of the lambs} & 4 & 0 \\
\rootlabel{0} & \textexneg{[UNK] of a movie} & 4 & 0 \\
\rootlabel{21} & \textexpos{ [UNK] [UNK] ,} & 0 & 4 \\
\rootlabel{70} & \textexpos{ daniel [UNK] } & 0 & 4 \\
\rootlabel{23} & \textexpos{year - old} & 0 & 4 \\
\rootlabel{22} & \textexpos{this " "} & 0 & 4 \\
\rootlabel{18} & \textexpos{\^ \& \# 214} & 0 & 4 \\
\rootlabel{0} & \textexpos{[UNK] of [UNK] [UNK]} & 0 & 4 \\
\rootlabel{11} & \textexpos{death of his father} & 0 & 3 \\
\rootlabel{26} & \textexneg{dickens '} & 3 & 0 \\
\bottomrule
\end{tabular}
}
\caption{\label{tab:additional_subj_features}
Additional SUBJ features (see Section~\ref{appendix:additional_features}).
We report the number of \ti{subjective} (\tf{S}) and \ti{objective} (\tf{O}) training examples associated with each feature and highlight the features according to the most common class.
These features reflect how this dataset was constructed~\citep{pang2004sentimental}: the \ti{subjective} class consists of movie reviews from Rotten Tomatoes and the \ti{objective} class consists of movie summaries from IMDb.
}
\end{table*}

%% file: tables/app_snli_additional_features.tex
\begin{table*}
\centering
\resizebox{\linewidth}{!}{%
\begin{tabular}{lp{\linewidth}rrr}
\toprule
\tf{Root} & \tf{Examples} & \textexpos{\tf{E}} & \textexneg{\tf{C}} & \textexneut{\tf{N}} \\
\midrule
\rootlabel{14} & \textexneg{a man/a woman}, \textexneg{a woman/a man}, \textexneg{a man/ nobody}, \textexneg{a boy/a girl}, \textexneg{a dog/a cat} & 235 & 1,690 & 307 \\
\rootlabel{17} & \textexneut{$\epsilon$/to work}, \textexneut{$\epsilon$/to get}, \textexneut{$\epsilon$/to buy}, \textexneut{$\epsilon$/the park}, \textexneut{$\epsilon$/on vacation} & 252 & 467 & 1,715 \\
\rootlabel{14} & \textexpos{a man/a person}, \textexpos{a man/a man}, \textexpos{a woman/a person}, \textexpos{ man/a man}, \textexpos{a man/a human} & 6,212 & 3,354 & 3,833 \\
\rootlabel{0} & \textexneg{in a/on the}, \textexneg{on a/in a}, \textexneg{on the/in the}, \textexneg{on a/in the}, \textexneg{on the/in a} & 555 & 1,810 & 793 \\
\rootlabel{4} & \textexpos{a /there is}, \textexpos{$\epsilon$/there are}, \textexpos{two /there are}, \textexpos{a /there are}, \textexpos{$\epsilon$/there is} & 1,356 & 449 & 402 \\
\rootlabel{0} & \textexpos{on a/on a}, \textexpos{at a/at a}, \textexpos{in a/in a}, \textexpos{on the/on the}, \textexpos{in a/wearing a} & 3,074 & 1,703 & 1,696 \\
\rootlabel{17} & \textexneg{$\epsilon$/at home}, \textexneg{$\epsilon$/in bed}, \textexneg{$\epsilon$/' t}, \textexneg{$\epsilon$/ice cream}, \textexneg{$\epsilon$/watching tv} & 56 & 597 & 210 \\
\rootlabel{2} & {\RaggedLeft \textexpos{in the grass/ outside}, \textexpos{down the street/ outside}, \textexpos{in the snow/ outside}, \textexpos{on the sidewalk/ outside}, \textexpos{in the snow/in the snow}} & 689 & 107 & 230 \\
\rootlabel{0} & \textexneut{on a/for a}, \textexneut{in a/for a}, \textexneut{on the/for a}, \textexneut{in the/ a}, \textexneut{of a/for a} & 157 & 222 & 752 \\
\rootlabel{14} & {\RaggedLeft \textexneut{a man/the man}, \textexneut{a man/a tall human}, \textexneut{a man/a couple}, \textexneut{a woman/the woman}, \textexneut{a woman/a tall human}} & 1,214 & 1,765 & 2,375 \\
\rootlabel{24} & \textexneg{black dog/ cat}, \textexneg{little girl/a boy}, \textexneg{young woman/a man}, \textexneg{ man/naked man}, \textexneg{brown dog/ cat} & 34 & 278 & 35 \\
\rootlabel{24} & \textexneut{ man/tall human}, \textexneut{ woman/tall human}, \textexneut{ man/old man}, \textexneut{ man/tall person}, \textexneut{ man/construction worker} & 83 & 135 & 368 \\
\rootlabel{24} & {\RaggedLeft \textexpos{group of people/there are people}, \textexpos{little boy/a boy}, \textexpos{black dog/ animal}, \textexpos{group of people/several people}, \textexpos{young boy/a child}} & 836 & 367 & 446 \\
\rootlabel{16} & \textexpos{in the/$\epsilon$}, \textexpos{on a/$\epsilon$}, \textexpos{down a/$\epsilon$}, \textexpos{through a/$\epsilon$}, \textexpos{through the/$\epsilon$} & 9,099 & 7,829 & 7,798 \\
\rootlabel{10} & \textexpos{towards the camera/$\epsilon$}, \textexpos{in the sand/$\epsilon$}, \textexpos{down a road/$\epsilon$}, \textexpos{with red - hair/$\epsilon$}, \textexpos{on the street/$\epsilon$} & 1,048 & 621 & 573 \\
\rootlabel{19} & {\RaggedLeft \textexpos{in the snow/ outside}, \textexpos{sitting on a bench/sitting on a bench}, \textexpos{in the snow/ outdoors}, \textexpos{in the air/ jumping}, \textexpos{in the grass/ outside}} & 78 & 2 & 5 \\
\rootlabel{21} & {\RaggedLeft \textexpos{in front/in front}, \textexpos{wearing hats/wearing hats}, \textexpos{wearing glasses/wearing glasses}, \textexpos{upside down/upside down}, \textexpos{wearing a hat/wearing a hat}} & 97 & 12 & 21 \\
\rootlabel{10} & \textexneg{in his hands/$\epsilon$}, \textexneg{' s hair/$\epsilon$}, \textexneg{wearing a brown/$\epsilon$}, \textexneg{with a bag/$\epsilon$}, \textexneg{on the road/$\epsilon$} & 28 & 86 & 20 \\
\rootlabel{11} & \textexpos{- hair/$\epsilon$}, \textexpos{green shirt/$\epsilon$}, \textexpos{striped shirt/$\epsilon$}, \textexpos{street vendor/$\epsilon$}, \textexpos{crowded street/$\epsilon$} & 160 & 81 & 64 \\
\rootlabel{5} & \textexpos{in front/$\epsilon$}, \textexpos{and shorts/$\epsilon$}, \textexpos{and smiling/$\epsilon$}, \textexpos{while people/$\epsilon$}, \textexpos{near water/$\epsilon$} & 572 & 396 & 383 \\
\rootlabel{4} & \textexneut{two young/the }, \textexneut{ two/the two}, \textexneut{four young/the }, \textexneut{$\epsilon$/one of}, \textexneut{$\epsilon$/a man and} & 91 & 147 & 204 \\
\rootlabel{2} & \textexneut{down the street/ home}, \textexneut{in the sand/on the beach}, \textexneut{at a table/ lunch}, \textexneut{on a bench/in a park} & 1 & 1 & 28 \\
\rootlabel{17} & \textexpos{$\epsilon$/at least}, \textexpos{$\epsilon$/a woman}, \textexpos{$\epsilon$/and child} & 39 & 4 & 14 \\
\rootlabel{16} & \textexneg{climb a/$\epsilon$}, \textexneg{playing a/$\epsilon$}, \textexneg{over their/$\epsilon$}, \textexneg{, "/$\epsilon$}, \textexneg{into the/$\epsilon$} & 249 & 304 & 180 \\
\rootlabel{11} & \textexneut{brown jacket/$\epsilon$}, \textexneut{blue plaid/$\epsilon$}, \textexneut{blue dress/$\epsilon$}, \textexneut{baseball uniform/$\epsilon$}, \textexneut{dark shirt/$\epsilon$} & 17 & 10 & 48 \\
\rootlabel{27} & \textexneg{of people/ person}, \textexneg{of people/ is}, \textexneg{of men/of women} & 28 & 81 & 48 \\
\rootlabel{31} & \textexpos{is walking/is walking}, \textexpos{is smiling/is smiling}, \textexpos{are dancing/are dancing} & 21 & 1 & 2 \\
\rootlabel{4} & \textexneg{two /there is}, \textexneg{two /there is only}, \textexneg{ three/the three}, \textexneg{three young/the }, \textexneg{two young/two old} & 21 & 67 & 38 \\
\rootlabel{27} & \textexpos{of people/are people}, \textexpos{of people/of people} & 210 & 138 & 129 \\
\rootlabel{8} & \textexneut{laughing /laughing at}, \textexneut{crying /crying because} & 0 & 0 & 14 \\
\rootlabel{10} & \textexneut{down a city/$\epsilon$}, \textexneut{in an office/$\epsilon$}, \textexneut{and hard hats/$\epsilon$}, \textexneut{wearing a black/$\epsilon$} & 31 & 20 & 60 \\
\rootlabel{27} & \textexneut{of people/of friends}, \textexneut{of dogs/ dogs} & 3 & 0 & 17 \\
\rootlabel{5} & \textexneg{wearing glasses/$\epsilon$}, \textexneg{around her/$\epsilon$}, \textexneg{in red/$\epsilon$}, \textexneg{of corn/$\epsilon$}, \textexneg{of volleyball/$\epsilon$} & 71 & 128 & 95 \\
\rootlabel{2} & \textexneg{at night/during the day} & 0 & 8 & 0 \\
\rootlabel{16} & \textexneut{on two/$\epsilon$}, \textexneut{in red/$\epsilon$}, \textexneut{and hard/$\epsilon$} & 56 & 67 & 92 \\
\rootlabel{5} & \textexneut{with black/$\epsilon$}, \textexneut{of young/$\epsilon$} & 38 & 28 & 54 \\
\rootlabel{11} & \textexneg{brown hat/$\epsilon$} & 0 & 7 & 1 \\
\rootlabel{15} & \textexpos{soccer ball/playing soccer} & 13 & 3 & 7 \\
\rootlabel{15} & \textexneut{tennis ball/playing fetch} & 0 & 0 & 5 \\
\rootlabel{6} & \textexneg{man and a/$\epsilon$} & 2 & 8 & 1 \\
\rootlabel{6} & \textexpos{man ' s/$\epsilon$} & 9 & 2 & 2 \\
\rootlabel{21} & \textexneut{playing soccer/playing soccer} & 11 & 4 & 14 \\
\rootlabel{6} & \textexneut{side of a/$\epsilon$} & 17 & 11 & 25 \\
\rootlabel{21} & \textexneg{hanging out/hanging out} & 0 & 6 & 2 \\
\bottomrule
\end{tabular}
}
\caption{\label{tab:additional_snli_features}
Additional SNLI features (see Section~\ref{appendix:additional_features}).
We report the number of \ti{entailment} (\tf{E}), \ti{contradiction} (\tf{C}), and \ti{netural} (\tf{N}) training examples associated with each feature and highlight the features according to the most common class.
\ti{Contradiction} features tend to involve antonyms, \ti{neutral} features tend to involve additions, and \ti{entailment} features involve copied clauses and hypernyms.
}
\end{table*}

%% file: tables/app_qqp_additional_features.tex
\begin{table*}
\centering
\resizebox{\linewidth}{!}{%
\begin{tabular}{lp{1.1\linewidth}rr}
\toprule
\tf{Root} & \tf{Examples} & \textexneg{\tf{N}} & \textexpos{\tf{P}} \\
\midrule
\rootlabel{25} & {\RaggedLeft{\textexpos{new year/new year}, \textexpos{world war/world war}, \textexpos{donald trump/donald trump}}} & 538 & 2,813 \\
\rootlabel{14} & {\RaggedLeft{\textexpos{how can/how can}, \textexpos{how do/how can}, \textexpos{how can/how do}}} & 4,072 & 7,952 \\
\rootlabel{31} & {\RaggedLeft{\textexpos{improve my/improve my}, \textexpos{earn money/earn money}, \textexpos{make money/make money}}} & 622 & 2,529 \\
\rootlabel{27} & {\RaggedLeft{\textexpos{candy imported/candy imported}, \textexpos{lose weight/lose weight}, \textexpos{writing skills/writing skills}}} & 93 & 1,257 \\
\rootlabel{14} & {\RaggedLeft{\textexneg{is /what is}, \textexneg{is /how do}, \textexneg{is /what are}}} & 1,759 & 410 \\
\rootlabel{3} & {\RaggedLeft{\textexpos{candy imported in/candy imported in}, \textexpos{not be/not be}, \textexpos{traffic to/traffic on}}} & 76 & 839 \\
\rootlabel{10} & {\RaggedLeft{\textexpos{i improve my/i improve my}, \textexpos{you have/you have}, \textexpos{i earn money/i earn money}}} & 316 & 1,336 \\
\rootlabel{7} & {\RaggedLeft{\textexpos{saltwater taffy/saltwater taffy}, \textexpos{way to/way to}, \textexpos{purpose of/purpose of}}} & 251 & 1,097 \\
\rootlabel{24} & {\RaggedLeft{\textexpos{do to/$\epsilon$}, \textexpos{to learn/to learn}, \textexpos{is //$\epsilon$}}} & 231 & 897 \\
\rootlabel{4} & {\RaggedLeft{\textexpos{$\epsilon$/do to}, \textexpos{$\epsilon$/way to}, \textexpos{$\epsilon$/and why}}} & 742 & 1,673 \\
\rootlabel{3} & {\RaggedLeft{\textexneg{[UNK] ./[UNK] .}, \textexneg{mean in/mean in}, \textexneg{politics and/politics and}}} & 349 & 16 \\
\rootlabel{2} & {\RaggedLeft{\textexneg{$\epsilon$/it like}, \textexneg{$\epsilon$/" "}, \textexneg{$\epsilon$/like to}}} & 642 & 154 \\
\rootlabel{4} & {\RaggedLeft{\textexneg{$\epsilon$/" "}, \textexneg{$\epsilon$/' t}, \textexneg{$\epsilon$/in [UNK]}}} & 740 & 218 \\
\rootlabel{24} & {\RaggedLeft{\textexneg{" "/$\epsilon$}, \textexneg{[UNK] in/$\epsilon$}, \textexneg{[UNK] [UNK]/$\epsilon$}}} & 558 & 127 \\
\rootlabel{25} & {\RaggedLeft{\textexneg{the word/the word}, \textexneg{the lewis/the lewis}, \textexneg{a sentence/a sentence}}} & 335 & 41 \\
\rootlabel{19} & {\RaggedLeft{\textexpos{hollywood movies/hollywood movies}, \textexpos{day of your life/day of your life}, \textexpos{company in delhi/company in delhi}}} & 5 & 158 \\
\rootlabel{27} & {\RaggedLeft{\textexneg{[UNK] . com/[UNK] . com}, \textexneg{politics and government/politics and government}, \textexneg{blood pressure/blood pressure}}} & 128 & 1 \\
\rootlabel{2} & {\RaggedLeft{\textexpos{$\epsilon$/? what are}, \textexpos{$\epsilon$/? what}, \textexpos{$\epsilon$/if yes}}} & 74 & 311 \\
\rootlabel{10} & {\RaggedLeft{\textexneg{the word `/the word `}, \textexneg{you determine the lewis/ is the lewis}, \textexneg{i watch/i watch}}} & 103 & 8 \\
\rootlabel{8} & {\RaggedLeft{\textexpos{$\epsilon$/? what are some examples}, \textexpos{$\epsilon$/from your perspective}, \textexpos{$\epsilon$/? how do they}}} & 18 & 108 \\
\rootlabel{19} & {\RaggedLeft{\textexneg{[UNK] . com/[UNK] . com}, \textexneg{college in singapore/college in singapore}}} & 52 & 0 \\
\rootlabel{18} &{\RaggedLeft{\textexpos{time travel to/time travel }, \textexpos{life ?/life }, \textexpos{spotify is/spotify }}} & 0 & 49 \\
\rootlabel{11} & {\RaggedLeft{\textexpos{will win/will win}, \textexpos{i can/ do}, \textexpos{music do/music do}}} & 8 & 75 \\
\rootlabel{5} & {\RaggedLeft{\textexpos{best day of your life/best day of your life}, \textexpos{purpose of life/purpose of life}, \textexpos{meaning of life/meaning of life}}} & 1 & 53 \\
\rootlabel{31} & {\RaggedLeft{\textexneg{solve this/solve this}, \textexneg{determine the lewis/is the lewis}, \textexneg{calculate the/calculate the}}} & 60 & 3 \\
\rootlabel{7} & {\RaggedLeft{\textexneg{review of/review of}, \textexneg{kind of/kind of}, \textexneg{aspects about/aspects about}}} & 515 & 303 \\
\rootlabel{8} & {\RaggedLeft{\textexneg{$\epsilon$/it like to}, \textexneg{$\epsilon$/it like to be}, \textexneg{$\epsilon$/like to be}}} & 43 & 1 \\
\rootlabel{11} & {\RaggedLeft{\textexneg{competitive is/competitive is}, \textexneg{much does/much does}, \textexneg{business can/business can}}} & 92 & 21 \\
\rootlabel{15} & {\RaggedLeft{\textexneg{what is [UNK] . com/what is [UNK] . com}, \textexneg{is [UNK] . com legit/is [UNK] . com legit}, \textexneg{what is the meaning of marathi word ` [UNK] '/what is the meaning of marathi word ` [UNK] '}}} & 38 & 0 \\
\rootlabel{9} & {\RaggedLeft{\textexneg{what is [UNK] . com ?/what is [UNK] . com ?}, \textexneg{is [UNK] . com legit ?/is [UNK] . com legit ?}, \textexneg{how is the word ` [UNK] ' used in a sentence ?/how is the word ` [UNK] ' used in a sentence ?}}} & 38 & 0 \\
\rootlabel{6} & {\RaggedLeft{\textexneg{[UNK] . com/[UNK] . com}, \textexneg{[UNK] [UNK]/[UNK] [UNK]}}} & 44 & 3 \\
\rootlabel{1} & {\RaggedLeft{\textexneg{is [UNK] . com/is [UNK] . com}, \textexneg{is [UNK] [UNK]/is [UNK] [UNK]}}} & 32 & 0 \\
\rootlabel{6} & {\RaggedLeft{\textexpos{the best day of your life/the best day of your life}, \textexpos{some examples/some examples}, \textexpos{the point of life/the point of life}}} & 3 & 31 \\
\rootlabel{0} & {\RaggedLeft{\textexpos{$\epsilon$/from your perspective ,}, \textexpos{$\epsilon$/we can remain satisfied in}, \textexpos{$\epsilon$/wanna ask someone please}}} & 0 & 21 \\
\rootlabel{12} & {\RaggedLeft{\textexneg{[UNK] . com make money/[UNK] . com make money}, \textexneg{[UNK] . com legit/[UNK] . com legit}, \textexneg{the word ` [UNK] ' used in a sentence/the word ` [UNK] ' used in a sentence}}} & 21 & 0 \\
\rootlabel{21} & {\RaggedLeft{\textexpos{long distance relationships/long distance relationship}, \textexpos{your life ? what happened/your life }, \textexpos{long distance relationships work/long distance relationship }}} & 0 & 19 \\
\rootlabel{20} & {\RaggedLeft{\textexpos{spotify is not/spotify }, \textexpos{new year ’ s/new year }, \textexpos{your life ? what/your life }}} & 0 & 18 \\
\rootlabel{26} & {\RaggedLeft{\textexneg{[UNK] /[UNK] and}, \textexneg{[UNK] /" "}}} & 18 & 1 \\
\rootlabel{13} & {\RaggedLeft{\textexpos{a person/i }}} & 2 & 16 \\
\rootlabel{0} & {\RaggedLeft{\textexneg{$\epsilon$/it like to be}}} & 9 & 0 \\
\rootlabel{15} & {\RaggedLeft{\textexpos{what are some examples/what are some examples}}} & 3 & 15 \\
\rootlabel{1} & {\RaggedLeft{\textexpos{are some examples/are some examples}}} & 3 & 15 \\
\rootlabel{12} & {\RaggedLeft{\textexpos{trump win/trump win}}} & 0 & 8 \\
\rootlabel{18} & {\RaggedLeft{\textexneg{[UNK] [UNK]/[UNK] [UNK]}}} & 25 & 10 \\
\rootlabel{13} & {\RaggedLeft{\textexneg{" "/it }}} & 7 & 0 \\
\rootlabel{5} & {\RaggedLeft{\textexneg{meaning of marathi word ` [UNK] '/meaning of marathi word ` [UNK] '}}} & 6 & 0 \\
\bottomrule
\end{tabular}
}
\caption{\label{tab:additional_qqp_features}
Additional QQP features (see Section~\ref{appendix:additional_features}).
We report the number of \ti{non-paraphrase} (\tf{N}) and \ti{paraphrase} (\tf{P}) training examples associated with each feature and highlight the features according to the most common class.
The \ti{paraphrase} features tend to correspond to how-to questions (such as how to earn money) or open-ended discussion questions---for example, about the 2016 presidential elections and the meaning of life.
The \ti{no paraphrase} features include subtrees reflecting sequences that appear in both questions and differ only in one, uncommon word, which is replaced with the unknown token. Training examples with this feature include ``What is instagramtop.com? What is bestmytest.com?'', or ``How is the word `wry' used in a sentence? How is the word `adduce' used in a sentence?'' This pattern appears exclusively in non-paraphrase examples.
}
\end{table*}

%% file: tables/snli_nonterminals.tex
\begin{table*}[ht]
\centering
\resizebox{\linewidth}{!}{%
\begin{tabular}{l p{0.85\linewidth} r r r }
\toprule
\tf{Production rule} & \tf{Spans} & \tf{E} & \tf{C} & \tf{N} \\
\midrule
\rootlabel{17} $\to$ \rootlabel{49} \rootlabel{35} & (17 (49 /to) (35 /work)) (17 (49 /to) (35 /get)) (17 (49 /to) (35 /buy)) & 1,990 & 3,774 & 6,397 \\
\rootlabel{2} $\to$ \rootlabel{17} \rootlabel{2} & (2 (17  /the toy) (2 between his legs/between his legs)) (2 (17  /a picture) (2 in the snow/in the snow)) (2 (17  /marco polo) (2 in the pool/in the pool)) & 627 & 1,495 & 2,415 \\
\rootlabel{2} $\to$ \rootlabel{0} \rootlabel{83} & (2 (0 at /during the) (83 night/day)) (2 (0 in the/in the) (83 snow/sand)) (2 (0 down a/down a) (83 street/street)) & 7,704 & 10,499 & 10,112 \\
\rootlabel{2} $\to$ \rootlabel{8} \rootlabel{17} & (2 (8  /outside in) (17  /the summer)) (2 (8  /off to) (17  /some friends)) (2 (8  /music on an) (17  /outside stage)) & 225 & 463 & 1,111 \\
\rootlabel{8} $\to$ \rootlabel{2} \rootlabel{49} & (8 (2 having a conversation/ talking) (49 /about)) (8 (2 an instrument/ music) (49 /for)) & 260 & 595 & 1,200 \\
\rootlabel{19} $\to$ \rootlabel{78} \rootlabel{2} & (19 (78 walking/walking) (2 down the street/in the mall)) (19 (78 jumping/sitting) (2 in the air/in a chair)) (19 (78 running/swimming) (2 through the snow/in a lake)) & 3,636 & 5,361 & 5,251 \\
\rootlabel{8} $\to$ \rootlabel{8} \rootlabel{49} & (8 (8 cheering /cheering for) (49 /their)) (8 (8 walking /walking down) (49 /a)) & 161 & 265 & 772 \\
\rootlabel{19} $\to$ \rootlabel{44} \rootlabel{10} & (19 (44 walking/walking) (10 down the street/ )) (19 (44 running/running) (10 through the water/ )) (19 (44 singing/singing) (10 into a microphone/ )) & 1,232 & 543 & 416 \\
\rootlabel{8} $\to$ \rootlabel{21} \rootlabel{49} & (8 (21 playing soccer/playing soccer) (49 /in)) (8 (21 taking a picture/taking a picture) (49 /of)) (8 (21 for a picture/for a picture) (49 /after)) & 260 & 361 & 949 \\
\rootlabel{2} $\to$ \rootlabel{49} \rootlabel{2} & (2 (49 /to) (2 on the corner/cross the street)) (2 (49 /while) (2 reading a book/ reading)) (2 (49 /to) (2 down the street/ work)) & 369 & 685 & 1,130 \\
\rootlabel{19} $\to$ \rootlabel{8} \rootlabel{17} & (19 (8 laughing /laughing at) (17  /a joke)) & 223 & 414 & 847 \\
\rootlabel{19} $\to$ \rootlabel{21} \rootlabel{10} & (19 (21 playing soccer/playing soccer) (10 on a field/ )) (19 (21 catching a football/catches a football) (10 with both hands/ )) (19 (21 be towed/being towed) (10 by a aaa/ )) & 1,690 & 1,004 & 797 \\
\rootlabel{15} $\to$ \rootlabel{92} \rootlabel{2} & (15 (92 cigarette/smoking) (2 in his mouth/ a pipe)) (15 (92 front/sitting) (2 of a building/ down)) & 957 & 1,943 & 1,627 \\
\rootlabel{4} $\to$ \rootlabel{89} \rootlabel{43} & (4 (89 a/there) (43 /is)) (4 (89 two/there) (43 /are)) (4 (89 a/there) (43 /are)) & 1,071 & 503 & 427 \\
\bottomrule
\end{tabular}
}
\caption{
\label{tab:snli_nonterminals}
The highest ranked binary production rules by mutual information in SNLI, grouped by majority class (\tf{Entailment}, \tf{Contradiction}, or \tf{Neutral}). Each row shows up to three of the highest scoring subtrees that are generated by that rule and have the same majority class label. %
}
\end{table*}

%% file: tables/snli_contrasting_example.tex
\begin{table*}[ht]
\centering
\small
\begin{tabular}{l p{0.4\linewidth} p{0.4\linewidth} r r}
  \toprule
  ID & Premise & Hypothesis & $y$ & $\hat{y}$ \\
  \midrule
  $\sx_0$ & a white dog running down a path & a black dog sitting on a bush & C & C \\
  \midrule
  $\sx_1$ & a white dog running down a path & a black dog running down a path & C & C \\
  $\sx_2$ & a white dog running down a path & a dog running down a path & E & E \\
  $\sx_3$ & a dog running down a white path & a dog running down a path & E & E \\
  $\sx_4$ & a dog running down a path & a black dog running down a path & N & N \\
  $\sx_5$ & a dog running down a white path & a black dog running down a path & N & C \\
  \bottomrule
\end{tabular}
\caption{
\label{tab:snli_contrasting_example}
A set of manual contrastive edits we create for the ``Adjective antonym'' feature. $y$ is the intended label (Entailment, Contradiction, or Neutral) and $\hat{y}$ is the prediction of a BERT classifier. See Section~\ref{appendix:additional_contrast sets}.
}
\end{table*}